\documentclass{article}
\usepackage[T1]{fontenc}
\usepackage{iclr2027_conference,times}
\usepackage{amsmath,amssymb,amsthm,booktabs,longtable,graphicx,float,microtype,xcolor}
\usepackage{adjustbox}
\usepackage{colortbl}
\definecolor{quantumrow}{RGB}{232,241,250}
\usepackage{caption}
\usepackage{hyperref}
\let\cite\citep
\usepackage{cleveref}
\crefname{equation}{Eq.}{Eqs.}
\Crefname{equation}{Eq.}{Eqs.}
\crefname{figure}{Figure}{Figures}
\Crefname{figure}{Figure}{Figures}
\crefname{table}{Table}{Tables}
\Crefname{table}{Table}{Tables}
\crefname{section}{Section}{Sections}
\Crefname{section}{Section}{Sections}
\crefname{subsection}{Section}{Sections}
\Crefname{subsection}{Section}{Sections}
\crefname{proposition}{Proposition}{Propositions}
\Crefname{proposition}{Proposition}{Propositions}
\crefname{appendix}{Appendix}{Appendices}
\Crefname{appendix}{Appendix}{Appendices}
\usepackage{placeins}
\newcommand{\method}{Quantum-JEPA}
\newcommand{\sg}{\operatorname{sg}}
\newcommand{\E}{\mathbb{E}}

\hypersetup{colorlinks=true,linkcolor=blue,citecolor=blue,urlcolor=blue}
\title{Quantum-JEPA: Quantum Born Priors\\
for Latent-Action JEPA World Models}
\author{Zhiyuan Liu$^{*,1,2}$ \quad Sicheng Liu$^{1}$ \quad Haiyu Wu$^{3}$\\
\normalfont $^{1}$Qinous Technology Co., Ltd., Beijing, China\\
\normalfont $^{2}$Tsinghua University, Beijing, China\\
\normalfont $^{3}$University of Notre Dame, Notre Dame, IN, USA\\
\normalfont\ttfamily $^{*}$liuzhiyu22@mails.tsinghua.edu.cn}
\iclrfinalcopy
\renewcommand{\headrulewidth}{0pt}
\date{}
\begin{document}
\maketitle
% The template's final mode implies acceptance; omit that header for this preprint.
\lhead{}

\begin{abstract}
A single observation can admit several plausible futures, making uncertainty
an essential part of world modelling. However, quantum generative models
remain underexplored for this task. We introduce \method{}, a latent-action
JEPA framework that learns distributions of future representations using a
conditional quantum Born prior. Circuit measurements generate binary latent
actions, which a shared predictor maps to alternative future representations.
We train the quantum prior through the quality and diversity of these
predictions using a sampled energy objective. Our model learns distinct continuations
on constructed multimodal tasks and captures latent-action correlations
that support coordinated multi-agent futures. On D3IL, a benchmark of diverse
human demonstrations, our trained models generate multiple obstacle-crossing
routes from a common starting point without observed motion.
On the controlled tasks, our classical and quantum prior
implementations improve prediction validity and distributional fidelity
over the evaluated variational baselines. Our quantum priors combine
competitive predictive quality with a parameter-efficient sampling structure:
they avoid an exponentially sized categorical output layer while supporting
correlated latent actions. We further validate finite-shot parameter-shift
training, including training with readout noise.
Unlike most prior work, which focuses on simulation, we also verify the feasibility
of inference with trained priors on a real QPU.
These results establish a
trainable quantum component for JEPA future prediction and a basis for
exploring quantum-enhanced world models at larger scales.
\end{abstract}

\begin{figure}[H]
\centering
\includegraphics[width=\textwidth]{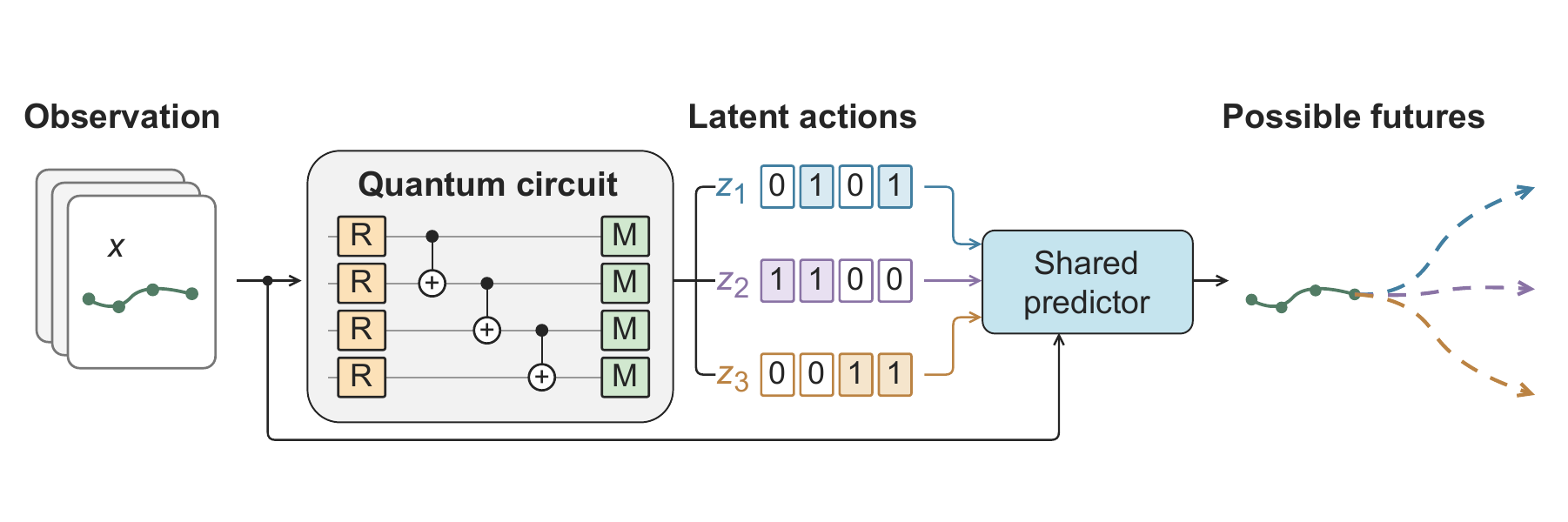}
\caption{Conditional quantum Born prior for latent-action generation.
The circuit samples binary actions conditioned on an observation; a shared
predictor maps these inputs to alternative future representations.
The solid trace depicts the observation, and colour-matched dashed branches
illustrate possible continuations.}
\label{fig:teaser}
\end{figure}

\section{Introduction}
LeCun's blueprint for autonomous machine intelligence assigns unpredictable
aspects of a future state to a latent variable within a representation-space
world model \cite{lecun2022path}. The latent variable can act as an unobserved
action: several continuations may be compatible with the same observation,
although only one is realized. Modelling this uncertainty requires both a
transition mechanism and a way to generate the missing information.
\Cref{fig:teaser} illustrates how we connect these two components through
quantum-sampled latent actions.

Joint-embedding predictive architectures (JEPAs) learn a transition by comparing
a prediction with an encoded target \cite{assran2023ijepa,bardes2024vjepa}.
A point-valued predictor supplies one representation for each observation.
Stochastic extensions instead describe a distribution of successor
representations \cite{huang2026varjepa,song2026branchjepa}. We study how to learn
this distribution through sampled latent actions. A useful prior must generate
valid alternatives, coordinate dependent changes and remain compact as the
action space grows. This motivates our joint study of the predictive-learning
mechanism and the quantum circuits that implement its generative prior.

Quantum machine learning explores how quantum representations and operations
can contribute to learning, including generative tasks in which sampling is
central \cite{biamonte2017qml}. We bring this perspective to JEPA by using a
conditional quantum circuit Born machine (QCBM) to generate latent actions.
Measuring an $n$-qubit register produces an $n$-bit code; one shared neural
predictor turns that code and the observation into a future representation.
We optimize the resulting predictive distribution with an energy score.
The observed future supplies a training signal for the conditional Born
distribution through the predictions generated from its samples.
We test both ideal-gradient optimization and parameter-shift gradients
estimated from finite measurement samples, including a readout-error channel
for our hardware-oriented reference circuits.
Encoding each action as an $n$-dimensional binary
vector keeps the predictor interface linear in register size while sharing
features across $2^n$ possible codes.

Quantum Born machines offer a direct connection between binary latent
variables and a physical sampling process. Product-QCBM circuits generate
conditionally independent bits, while entangling gates allow correlated
measurements \cite{liu2018qcbm,benedetti2019qcbm}. This provides a circuit-level
way to encode correlations between latent actions. Classical autoregressive
models and mixtures also generate correlations, and a shared neural predictor
can create output dependence even from independent random inputs.
We therefore restrict each local predictor to its own action bits in the
multi-agent experiment, isolating the prior's role in coordinating futures.
Our comparison considers predictive quality, consistency across training
runs and prior resources together, since each captures a different
requirement of a useful world model.

We evaluate the framework on stochastic MovingMNIST,
HighwayEnv-Priority18 (HEP18), a controlled four-agent intersection with
18 coordinated futures, and public D3IL human demonstrations
\cite{jia2024d3il}. The controlled tasks reveal mode recovery and coordination;
D3IL tests learning from individual observed continuations and recursively
predicting alternative routes from the same starting point. Comparisons with variational learning methods
assess the predictive mechanism, while classical prior comparisons isolate
the quantum implementation. Our experiments test whether sampled actions
produce valid image continuations and coordinated motion, and whether
those predictions retain the frequencies of alternative futures.
Measurement-based training in simulation and real-QPU inference
experiments test the two stages of this hybrid workflow.
Our resource analysis compares the parameters, sampling operations and
storage required by each prior. At fixed circuit depth, the QCBM uses
linearly many conditional angles and local quantum gates rather than
one logit for each joint code. On D3IL its prior uses fewer parameters
than AR-GRU and Softmax-Categorical.

Our contributions are:
\begin{enumerate}
\item A sampled latent-action JEPA framework that jointly learns future
representations and a conditional generative prior, improving sample validity
and distributional fidelity over the evaluated variational baselines on our
controlled tasks.
\item A trainable quantum Born implementation that learns alternative futures
and useful cross-agent correlations through finite-shot parameter-shift
training, together with deployment of trained priors on a QPU.
\item A performance--resource study spanning controlled multimodal tasks
and public demonstrations, with strong autoregressive controls, recursive
prediction and task-specific simplification checks. Our quantum priors
offer a competitive quality--parameter trade-off through circuit-based
sampling with a shared latent-action predictor.
\end{enumerate}

\section{Related Work}
\paragraph{Stochastic predictive world models.}
Stochastic video models use learned latent variables to represent alternative
continuations \cite{denton2018stochastic,babaeizadeh2018stochastic}.
JEPA moves prediction into representation space: I-JEPA and V-JEPA learn
by predicting target embeddings \cite{assran2023ijepa,bardes2024vjepa}.
Within this setting, Variational JEPA learns a predictive distribution through
a variational formulation \cite{huang2026varjepa}, while Branch-JEPA represents
a context-weighted finite set of future embeddings and trains it with
specialization or distributional scores \cite{song2026branchjepa}.
FlowWM models stochastic futures directly in frozen pretrained visual
features using conditional flow matching \cite{porcher2026flowwm}.
In contrast, we jointly train the JEPA encoder, shared predictor and
discrete latent-action prior through a sampled predictive energy objective.
Together, these works establish representation-space uncertainty modelling
as a broader direction beyond point prediction.
Our focus is how a sampled binary prior can learn through its consequences
in a shared latent transition model. We separate two questions experimentally:
how our sampled predictive-learning framework compares with other stochastic
learning objectives, and how quantum and classical priors compare within
that same framework.

\paragraph{Latent actions.}
Action-conditioned world models predict the consequences of a supplied control.
V-JEPA~2-AC extends pretrained predictive representations to this setting
\cite{assran2025vjepa2}; LeWorldModel (LeWM) learns latent dynamics from
observation--action trajectories using prediction and SIGReg losses
\cite{maes2026lewm}. When action labels are unavailable, latent-action models
instead learn a control interface from observed changes. VQ-VAE provides
a discrete representation through vector quantization \cite{oord2017vqvae},
and LAPA and Genie learn action-like variables from videos without action labels
\cite{ye2024lapa,bruce2024genie}. Work on latent-action world models in the wild
also studies constrained continuous actions and transfer across videos
\cite{garrido2026latentaction}. We use latent actions to index possible
continuations of the same observation. Their conditional distribution is
learned jointly with predictive representations, and binary codes provide
a direct interface to quantum measurements. Our intervention experiments
test whether changing this interface redirects the learned continuation.

\paragraph{Sample-based distribution learning.}
Proper scoring rules evaluate a predictive distribution against an observed
outcome \cite{gneiting2007proper}. Energy scores are useful when a
generator can sample but cannot conveniently evaluate its likelihood.
Scoring-rule minimization has been developed for neural probabilistic
forecasting \cite{pacchiardi2024forecasting}. We use this sample-based interface
to connect a generative prior to the JEPA prediction task: the score assesses
the future embeddings produced from sampled actions and supplies the prior's
learning signal. Since the target space is learned, a predictive score alone
does not prevent representation collapse. VICReg, LeJEPA and VISReg address
this separate problem through feature regularization
\cite{bardes2022vicreg,balestriero2025lejepa,wu2026visreg}.
Our framework combines a sampled predictive objective with VISReg; its
quantum instance estimates prior gradients through circuit executions.

\paragraph{Quantum generative models.}
Quantum circuit Born machines parameterize discrete probability distributions
with trainable quantum circuits, drawing binary samples by measuring the
states they prepare \cite{liu2018qcbm,benedetti2019qcbm}.
Conditional versions use classical side information to adapt the circuit's
output distribution to a given input \cite{kiss2022conditional}.
Parameter-shift rules make derivatives accessible through shifted circuit
executions \cite{schuld2019gradients}. Their usefulness depends on the loss
and ansatz: quantum generative modelling can encounter trainability barriers,
including weak statistical signals in high-dimensional objectives
\cite{rudolph2024trainability}.
HQ-JEPA uses a quantum fidelity loss for cross-modal remote-sensing
representation learning, with its quantum module evaluated through classical
simulation \cite{hossain2026hqjepa}. We instead place the circuit in the
generative prior, where it samples latent actions and learns through their
downstream predictive energy. This ties quantum distribution learning to
future uncertainty in a learned world model. We also deploy trained priors
on a real QPU and pass their measured actions through the complete prediction
pipeline. This establishes inference feasibility beyond simulation and
complements our classical-prior comparisons and correlation tests.

\section{Method}
\label{sec:method}
\subsection{Observation, latent action, and predictive energy}
\begin{figure}[tbp]
\centering
\includegraphics[width=\textwidth]{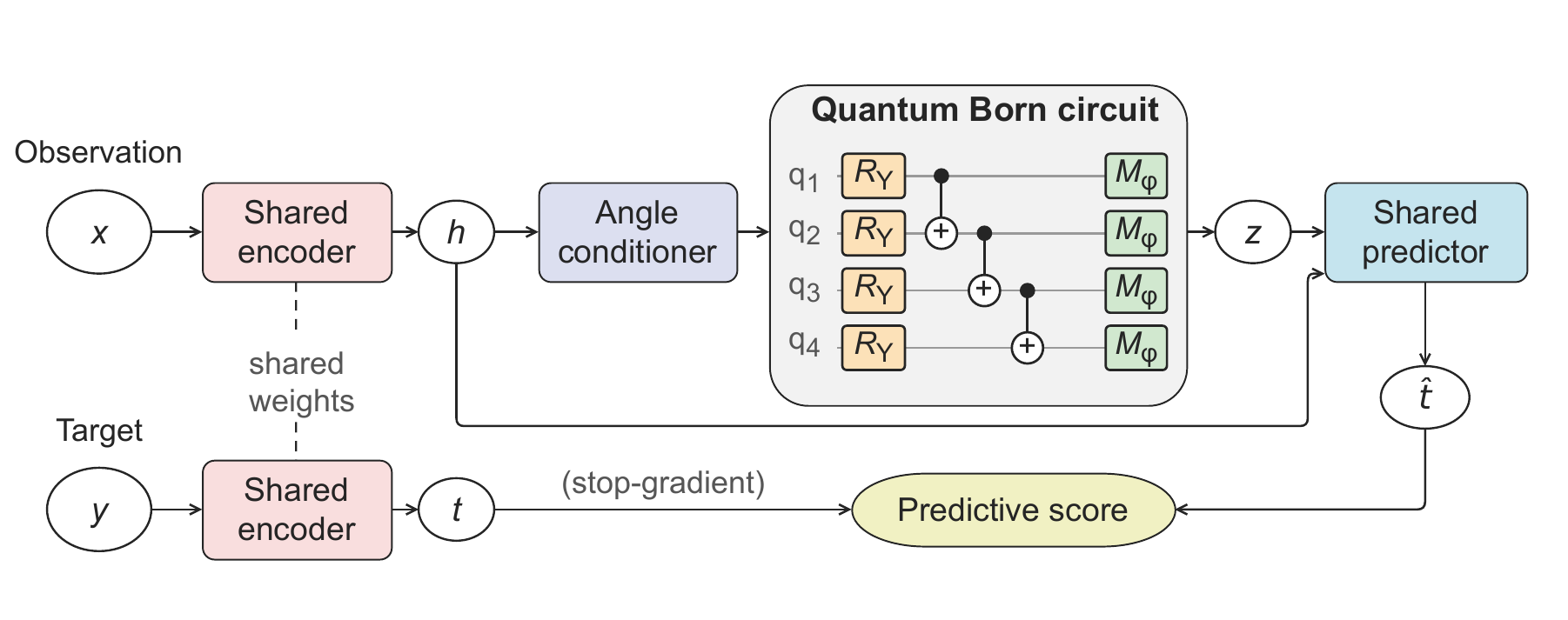}
\caption{Quantum-JEPA architecture. Filled blocks denote modules and
outlined tokens denote data. The observation representation $h$
conditions both the Born prior and the shared predictor. Measured bits $z$
select a predicted future representation $\widehat t$, which is scored
against the encoded target $t$. This score trains the predictor and prior;
feature regularization stabilizes the shared encoder and is omitted for
clarity. The circuit is schematic; $M_\varphi$ denotes the measurement stage.}
\label{fig:structure}
\end{figure}

Let $x$ denote an observation and $y$ one realized future. A shared encoder
$f_\xi$ maps the two views to $d$-dimensional representations,
\begin{equation}
 h=f_\xi(x),\qquad t=f_\xi(y).
 \label{eq:views}
\end{equation}
A prior $p_\theta(z\mid h)$ generates $z\in\{0,1\}^n$, and a shared predictor
$P_\psi$ produces
\begin{equation}
 Z\mid x\sim p_\theta(\cdot\mid f_\xi(x)),\qquad
 \widehat T=P_\psi(f_\xi(x),Z),\qquad
 \widehat T\mid x\sim Q_{\xi,\psi,\theta}(\cdot\mid x).
 \label{eq:pushforward}
\end{equation}
Here $\widehat T$ is the random future representation; a sampled code $z$
gives one prediction $\widehat t=P_\psi(h,z)$.
We use the latent action to represent transition information that remains
uncertain given the observation. The model learns its effect by comparing sampled
predictions with the target representation. Within the predictive-energy
gradient, we hold the target fixed so that this update improves the prediction
towards a reference representation. We train both encoder views through
representation regularization.
\Cref{fig:structure} summarizes the training paths.

\subsection{Conditional Born prior and shared code predictor}
We use several circuit structures with the same Born-sampling interface.
Our hardware-oriented RY/CNOT reference uses a conditioner
$g_\eta:\mathbb R^d\rightarrow\mathbb R^{L\times n}$, which
sets the angles of an $L$-layer conditional QCBM. Our ansatz comprises
rotation--entangling layers and a final layer of single-qubit rotations:
\begin{align}
 \alpha_{\ell j}(h)&=g_\eta(h)_{\ell j}+\vartheta_{\ell j}
       +\tfrac{\pi}{2}\mathbf 1_{\ell=1},\\
 U_{\alpha,\varphi}
 &=\left[\bigotimes_{j=1}^{n}R_Y(\varphi_j)\right]
   \prod_{\ell=1}^{L}\left[E_\ell\bigotimes_{j=1}^{n}R_Y(\alpha_{\ell j})\right],
 \qquad
 p_\theta(z\mid h)=|\langle z|U_{\alpha,\varphi}|0^n\rangle|^2.
 \label{eq:born}
\end{align}
The body layers act in temporal order, followed by the terminal rotations
and computational-basis measurement; $\theta=(\eta,\vartheta,\varphi)$.
Each $E_\ell$ is a fixed CNOT matching, or the identity for the product
circuit. The conditioner is a two-layer SiLU MLP whose output layer starts
at zero. Small offsets around the first-layer $\pi/2$ rotation provide broad
initial support. We initialize the $n$ terminal angles at zero and share
them across observations.

The final $R_Y(\varphi_j)$ rotations provide a trainable local readout
basis for the binary actions, without increasing the conditioner output
dimension. The prior generates bit strings by measurement, and the predictor
maps each string to a future representation. We estimate the prior gradient
from the resulting predictive costs. The measurement-based update does not
require a table of code probabilities.

The predictor embeds the raw signed bits with a two-layer MLP,
\begin{equation}
 u(z)=\sigma\!\left(W_{z2}\sigma(W_{z1}(2z-\mathbf1)+b_{z1})+b_{z2}\right),
 \quad c(h)=W_hh+b_h ,
 \label{eq:code}
\end{equation}
where $\sigma$ is SiLU. A residual transition then combines context, action,
and their elementwise interaction:
\begin{equation}
 P_\psi(h,z)=\mathcal N
 \left(h+W_2\sigma\!\left(W_1[c(h);u(z);c(h)\odot u(z)]+b_1\right)+b_2\right).
 \label{eq:predictor}
\end{equation}
Here $\mathcal N$ is an output normalization map, specified with each
benchmark architecture in Appendix~\ref{app:networks}. The input width
grows linearly with $n$, sharing parameters across all joint codes.
The D3IL implementation also learns a bounded output scale and offset,
described in Appendix~\ref{app:d3il-implementation}.

\subsection{Sampled predictive objective}
For $K\geq2$ independent prior samples, write
$\widehat t_k=P_\psi(h,z_k)$ and $\rho(u,v)=\|u-v\|_2/\sqrt d$.
We use the empirical energy loss
\begin{equation}
 \mathcal L_{\rm pred}
 =\frac1K\sum_{k=1}^{K}\rho(\widehat t_k,t)
 -\frac{1}{2K(K-1)}\sum_{k\ne l}\rho(\widehat t_k,\widehat t_l).
 \label{eq:energy}
\end{equation}
The first term brings predictions closer to the observed target.
The second encourages diversity among predictions, balancing target accuracy
with coverage of different possible futures.
Excluding self-pairs gives an unbiased estimate of the population energy
score for a fixed representation and independent draws
\cite{gneiting2007proper}.
When several realized futures share an observation, we average the first
term over those targets. The prior still depends only on the observation.

The encoder could reduce the predictive score simply by making its features
uninformative. We therefore use VISReg \cite{wu2026visreg} to maintain feature
scale and spread while learning the predictive task.
We use its decoupled centre, scale and shape components,
\begin{equation}
 R(v)=\lambda_{\rm center}\mathcal L_{\rm center}(v)
     +\lambda_{\rm scale}\mathcal L_{\rm scale}(v)
     +\lambda_{\rm shape}\mathcal L_{\rm shape}(v).
 \label{eq:regularizer}
\end{equation}
These terms centre features, control their coordinate-wise scale, and match
normalized random projections to Gaussian quantiles.
Appendix~\ref{app:visreg} specifies their implementation and weights.
We regularize the observation and target representations:
\begin{equation}
 \mathcal L_{\rm reg}=\frac{R(h)+R(t)}2.
 \label{eq:regviews}
\end{equation}
Our world-model objective is
\begin{equation}
 \boxed{\mathcal L_{\rm JEPA}
       =\mathcal L_{\rm pred}+\lambda_{\rm reg}\mathcal L_{\rm reg}.}
 \label{eq:loss}
\end{equation}
For repeated target views, we average $R(t)$ over views before combining
the two terms.

\subsection{Training latent priors from predictive quality}
\label{sec:prior-training}
We update the encoder and predictor by backpropagation with sampled codes
held fixed. We train the prior through the distribution of its predictions.
For a fixed context, target and predictor, the first variation of the
population predictive loss at prediction $u$ is
\begin{equation}
 C(u;t,Q)=\rho(u,t)-\E_{V\sim Q}\rho(u,V).
 \label{eq:firstvariation}
\end{equation}
The second term has coefficient one because both draws in the pairwise
energy depend on the prior. This cost rewards target fit together with
useful predictive diversity. We average its first term over target views
when several realized futures share an observation.

\paragraph{Classical prior update.}
For a classical prior, we use a leave-one-out estimate of this cost and
differentiate the log probability of each sampled code:
\begin{align}
 C_k^{\rm LOO}
 &=\rho(\widehat t_k,t)-\frac1{K-1}\sum_{l\ne k}\rho(\widehat t_k,\widehat t_l),\\
 \mathcal L_{\rm prior}^{\rm sur}
 &=\frac1K\sum_k\log p_\theta(z_k\mid\sg(h))\,
                      \sg(C_k^{\rm LOO}).
 \label{eq:prior-surrogate}
\end{align}
Here $\sg$ stops gradients, restricting this update to the prior and
favouring codes with lower predictive cost.
We average the target term over available futures; classical priors supply
their respective code log probabilities.

\paragraph{Measurement-based quantum update.}
Let $a$ collect the body and terminal $R_Y$ rotation angles of the
conditional QCBM. For each angle $a_j$, the
parameter-shift rule \cite{schuld2019gradients} gives
\begin{equation}
 \frac{\partial\overline{\mathcal L}_{\rm pred}}{\partial a_j}
 =\frac12\left[
 \E_{Z\sim p_{a+\frac\pi2e_j}} C(P_\psi(h,Z);t,Q)
 -\E_{Z\sim p_{a-\frac\pi2e_j}} C(P_\psi(h,Z);t,Q)
 \right].
 \label{eq:shift}
\end{equation}
We estimate the two expectations with shifted-circuit measurements and
use independent unshifted predictions as references for $Q$. The target,
predictor and reference distribution are fixed during this prior update.
The estimated angle derivatives are backpropagated through the classical
conditioner, which receives $\sg(h)$. This trains the Born prior through
its downstream predictions without differentiating a quantum statevector.

We subsample observations and angle coordinates with inverse-probability
corrections to control measurement cost. The rule remains valid with a
parameter-independent readout channel, which we apply to shifted and
unshifted samples in the RY/CNOT reference experiments. Ideal-gradient quantum experiments
use \cref{eq:prior-surrogate} with differentiable Born log probabilities
(Appendix~\ref{app:gradient}). All priors share the predictive objective and $K$
prediction samples; quantum gradients require additional shots.
Appendix~\ref{app:training} gives their budgets and the warm-up schedule.

\section{Correlated Latent Priors and Resource Scaling}
\subsection{Prior families}
We compare explicit categorical, independent-bit, autoregressive, mixture
and quantum priors for the same $n$-bit action space.
Their parameterizations trade off joint-distribution capacity, parameter
count and sampling work. \Cref{sec:comparison-methods} defines each method;
Appendix~\ref{app:training} gives its implementation.

\subsection{Resource comparison}
\label{sec:resource-crossover}
Let $H_p$ be the prior's hidden width, $L$ the circuit depth in layers, and
$R$ the number of mixture components. \Cref{tab:resources} excludes the
encoder and predictor, which are shared across comparison methods.
The classical conditioner of the Born prior is part of its cost.

\begin{table}[tbp]
\centering\normalsize
\setlength{\tabcolsep}{3pt}
\caption{Leading prior costs for one observation and $K$ sampled $n$-bit
codes. Parameter counts are asymptotic; constants, biases, and normalization
parameters are omitted. Classical costs allow cached context projections
and alias-table categorical sampling. Quantum gate counts assume compatible connectivity.
Every method also stores $O(Kn)$ measured or sampled bits.}
\label{tab:resources}
\begin{tabular}{@{}lll@{}}
\toprule
Prior & Parameters & Generation work \\
\midrule
Softmax-Categorical & $O(dH_p+H_p2^n)$ & $O(dH_p+H_p2^n+K n)$ \\
Independent-Bernoulli & $O(dH_p+H_p n)$ & $O(dH_p+H_p n+Kn)$ \\
Mixture-$R$ & $O(dH_p+H_p Rn)$ &
 $O(dH_p+H_p Rn+R+Kn)$ \\
AR-GRU & $O(dH_p+H_p^2)$ & $O(dH_p+KnH_p^2)$ \\
\midrule
Conditional QCBM & $O(dH_p+H_p nL)$ &
 \begin{tabular}[c]{@{}l@{}}$O(dH_p+H_p nL)$ classical\\$O(KnL)$ quantum gates\end{tabular} \\
\bottomrule
\end{tabular}
\end{table}

At fixed depth and conditioner width, QCBM parameter and gate counts grow
linearly with $n$, without a $2^n$-entry output layer. Independent bits are
also compact but factorized; mixtures introduce a component-count cost,
and AR-GRU samples bits sequentially. A QCBM produces correlated bits in
one circuit execution per draw, with correlations determined by its gates
and depth. Appendix~\ref{app:resource-details} gives working-memory assumptions.

\section{Experimental Protocol}
\subsection{Common training and comparison methods}
\label{sec:comparison-methods}
Within each experiment, stochastic discrete-prior comparisons share the encoder, predictor, loss,
readout and training schedule.
Deterministic JEPA predicts one future per observation; Uniform-Bernoulli
uses independent fair bits, and Frozen-QCBM keeps the initialized prior fixed.
Softmax-Categorical learns one probability per code, whereas
Independent-Bernoulli learns a separate probability for each bit.
AR-GRU conditions each bit on preceding bits; a Bernoulli mixture first
samples a shared component and then draws conditionally independent bits.
Product-QCBM uses single-qubit rotations, while Entangled-QCBM
adds CNOT gates between them to couple the sampled bits.
Mixture-2 for MovingMNIST and Mixture-4 for HEP18 and D3IL have
parameter counts close to the corresponding QCBM.
For complete-method comparisons, our Variational JEPA adaptation
\cite{huang2026varjepa} predicts a mean representation and calibrates
Gaussian uncertainty.
Latent-CVAE \cite{denton2018stochastic,babaeizadeh2018stochastic} learns
a future-conditioned posterior and context-conditioned prior with a
reconstruction--KL objective. Both use their own objectives with matched
data, readouts and evaluation budgets.
The reference results in \cref{tab:mnist,tab:intersection} use three
independent trainings and 128 actual prior draws per observation.
Their trained QCBMs use finite-shot parameter
shift with independent 0.5\% readout bit flips during training and
evaluation; classical methods and fixed controls retain their stated
sampling protocols. Appendix~\ref{app:gradient-comparison} compares this
setting with ideal gradients and noise-free parameter shift.
D3IL training covers three seeds across three demonstration folds;
the common-start diagnostic evaluates all nine models per method.
Its trainable QCBMs also use finite-shot parameter shift with 0.5\% readout
bit flips; the frozen quantum control receives the same readout noise.
We report each experiment separately, without pooling
checkpoints across circuit families, gradient estimators or data splits.
Architectures, optimizers and external adaptations appear in
Appendices~\ref{app:training}, \ref{app:circuits} and \ref{app:external}.

\subsection{Stochastic MovingMNIST}
\label{sec:mnist}
We construct a stochastic single-digit variant of MovingMNIST
\cite{srivastava2015unsupervised} using $64\times64$ videos
with three horizontal-motion observation frames and three future frames.
Eight equally spaced directions have probabilities 0.15 for each cardinal
direction and 0.10 for each diagonal direction. We use 800 training and
80 test observations from disjoint MNIST splits. Training receives image
pairs, not motion labels; a three-bit prior learns the action codes.
These frequencies admit a product-bit representation, so this task does
not require entanglement. We train with \cref{eq:loss}, using VISReg on
the observation and target representations. A separately trained frozen-JEPA readout predicts
translations and renders future images from the last observation frame.
Appendix~\ref{app:mnist} gives the full construction and readout protocol.

\subsection{HighwayEnv priority intersection (HEP18)}
\label{sec:intersection}
HEP18 is a controlled four-agent task on HighwayEnv's straight intersection
lanes \cite{leurent2018highwayenv}. Each observation admits 18 equally likely
futures: either NS or EW has priority, each priority vehicle chooses one
of three speed schedules, and both yielding vehicles decelerate to a stop.
The reference experiment splits 448 observations into 256 training,
64 validation and 128 test contexts, keeping each observation's futures
together. We train the JEPA for 6,000 updates and fit the readout for
16,000 updates. All 18 futures are available during HEP18 training;
their probabilities are uniform and do not depend on the initial state.
The model consumes structured position--speed histories and predicts
three stages over nine seconds.

We use twelve bits, three per agent. Each local predictor sees the full
deterministic context but only its own action bits, so stochastic
coordination must come from the prior. Training scores the concatenated
scene with the smoothed power distance in \cref{eq:power-distance} and
adds $\lambda_\Delta R(t-h)$ to preserve change information.
A frozen per-agent readout predicts speeds for kinematic integration.
Appendix~\ref{app:intersection} specifies these task-specific choices;
Appendix~\ref{app:objective-stability} tests change regularization.

\subsection{D3IL demonstration continuation}
\label{sec:d3il}
We use 96 public human demonstrations from D3IL Avoiding
\cite{jia2024d3il}, a simulated robotic obstacle-avoidance task with three obstacle rows.
We train the models on observed planar trajectories, without enumerating alternative
futures for each observation or supplying route labels to the model.
We train on demonstration-disjoint folds of 64 trajectories, reserving
32 in each fold, with three training seeds and eight methods.
For the common-start evaluation below, we evaluate all nine trained models
per method. We average the three folds within each training seed, then report
the mean and sample standard deviation of the three seed-level means.
Figures use the first training seed and first fold for every method,
without selecting checkpoints by performance.
The dataset informed earlier development, and this diagnostic is not
a previously untouched blind test.

An observation is a 24-point trajectory segment. We predict the next
8\,cm of motion, also represented by 24 points, using eight latent-action
bits. After decoding a prediction, we re-encode its coordinates for the
next step. We stop at the goal line or after 16 calls, retaining failures
in evaluation. This is offline recursive prediction, not robot execution.
Training uses the Euclidean energy score and VISReg on $h$ and $t$.
Here Entangled-QCBM uses four layers of local rotations and brickwork
CNOT gates, followed by terminal local rotations; Product-QCBM omits the CNOTs.
Frozen-QCBM retains its input-independent initialized distribution
while the encoder and predictor learn. Deterministic JEPA uses representation
MSE in place of the energy score.
AR-GRU uses the same recurrent prior architecture as in the other two tasks.
A frozen-JEPA spline readout translates future features into coordinates.
We retain a final 1,000-update predictor/prior stage on real consecutive
segments, with encoder and readout fixed. Appendix~\ref{app:d3il-implementation}
gives all stages and architecture differences, including the CVAE schedule.

All 96 recorded trajectories begin at exactly the same planar position.
We use that point as the entire initial observation, repeating it 24 times
to retain the encoder's input shape without supplying a motion direction.
Subsequent calls re-encode the predicted segment as usual.
During training, the first observation is a moving segment interpolated
from recorded frames 4--8.
At evaluation, we keep the trained models fixed and begin prediction
at the shared stationary point, before any motion has been observed.

\subsection{Prediction quality and coordination metrics}
\label{sec:metrics}
Coverage measures how many true futures are recovered; validity measures
the fraction of draws matching a true future. Total variation (TV)
compares sampled mode frequencies, including an invalid category, with
the target distribution. MovingMNIST uses a 4-pixel endpoint threshold.
HEP18 requires every agent to match the same joint target within 2\,m
and 1\,m/s at all future times.
We also report minimum ADE and Euclidean trajectory energy score (ES).
Independently re-pairing agents'
predictions preserves their marginals but disrupts coordination.
Appendix~\ref{app:metric-definitions} gives complete definitions and
Appendix~\ref{app:metrics} reports tolerance and sampling-budget checks.

For D3IL, the three obstacle rows define $2\times3\times4=24$ routes.
We draw 384 recursive continuations per checkpoint from the same point.
A geometrically accepted path reaches the goal, crosses all three rows,
avoids continuous-path collisions and passes a workspace-envelope check.
We report the exact expected route coverage and TV of a uniform random
subset of 60 accepted paths, against the uniform 24-route reference.
One Latent-CVAE checkpoint supplies only 61 accepted paths; the common
60-path budget retains every stochastic checkpoint without additional draws.
Unlike the controlled-task TV above,
this TV is conditioned on geometric validity; it does not measure the
probability of producing an invalid path. All raw candidates are retained.
Appendix~\ref{app:d3il-metrics} gives the definitions, the empirical
demonstration reference and a finite-pool sampling-sensitivity check.

\section{Results}
\subsection{Multimodal prediction on MovingMNIST}
\begin{table}[tbp]
\centering\normalsize
\setlength{\tabcolsep}{2pt}
\medmuskip=0mu
\caption{MovingMNIST (80 observations, 128 draws each). Mean $\pm$ sample s.d. over three trainings; coverage/validity in percent, minADE/ES in pixels. The first two rows are external methods. Trained QCBMs use parameter shift with 0.5\% readout flips. Shading marks trained QCBMs; bold/underline mark best/second-best displayed means, including ties.}
\label{tab:mnist}
\begin{tabular}{@{}lrrrrr@{}}
\toprule
Method / prior & \multicolumn{1}{c}{Cov. $\uparrow$} & \multicolumn{1}{c}{Valid $\uparrow$} & \multicolumn{1}{c}{TV $\downarrow$} & \multicolumn{1}{c}{minADE $\downarrow$} & \multicolumn{1}{c}{ES $\downarrow$} \\
\midrule
Variational JEPA & $63.3\,\pm\,8.1$ & $31.0\,\pm\,2.3$ & $0.730\,\pm\,0.009$ & $2.43\,\pm\,0.30$ & $11.55\,\pm\,0.66$ \\
Latent-CVAE & $\mathbf{98.6}\,\pm\,1.0$ & $53.5\,\pm\,2.3$ & $0.470\,\pm\,0.023$ & $\mathbf{0.70}\,\pm\,0.03$ & $9.96\,\pm\,0.05$ \\
\midrule % external-method separator
Deterministic JEPA & $0.0\,\pm\,0.0$ & $0.0\,\pm\,0.0$ & $1.000\,\pm\,0.000$ & $8.04\,\pm\,0.01$ & $15.04\,\pm\,0.01$ \\
Uniform-Bernoulli & $\underline{94.4}\,\pm\,1.6$ & $\mathbf{96.5}\,\pm\,1.2$ & $0.168\,\pm\,0.018$ & $0.81\,\pm\,0.08$ & $\underline{9.60}\,\pm\,0.02$ \\
Softmax-Categorical & $93.6\,\pm\,0.5$ & $95.8\,\pm\,0.5$ & $\mathbf{0.145}\,\pm\,0.002$ & $\underline{0.79}\,\pm\,0.04$ & $\mathbf{9.58}\,\pm\,0.02$ \\
AR-GRU & $90.5\,\pm\,4.1$ & $96.0\,\pm\,1.1$ & $0.180\,\pm\,0.015$ & $0.97\,\pm\,0.18$ & $9.65\,\pm\,0.06$ \\
Mixture-2 & $88.5\,\pm\,4.4$ & $95.8\,\pm\,1.2$ & $0.186\,\pm\,0.020$ & $0.98\,\pm\,0.19$ & $9.63\,\pm\,0.05$ \\
\midrule % prior-family separator
Frozen-QCBM & $93.3\,\pm\,3.7$ & $95.6\,\pm\,1.8$ & $0.176\,\pm\,0.027$ & $0.84\,\pm\,0.13$ & $9.61\,\pm\,0.04$ \\
\rowcolor{quantumrow}
Product-QCBM & $94.3\,\pm\,2.4$ & $\underline{96.2}\,\pm\,1.6$ & $0.167\,\pm\,0.019$ & $0.80\,\pm\,0.08$ & $\underline{9.60}\,\pm\,0.02$ \\
\rowcolor{quantumrow}
Entangled-QCBM & $94.0\,\pm\,2.3$ & $\mathbf{96.5}\,\pm\,1.1$ & $\underline{0.160}\,\pm\,0.024$ & $0.84\,\pm\,0.11$ & $9.61\,\pm\,0.03$ \\
\bottomrule
\end{tabular}
\end{table}

\begin{figure}[tbp]
\centering
\includegraphics[width=\textwidth]{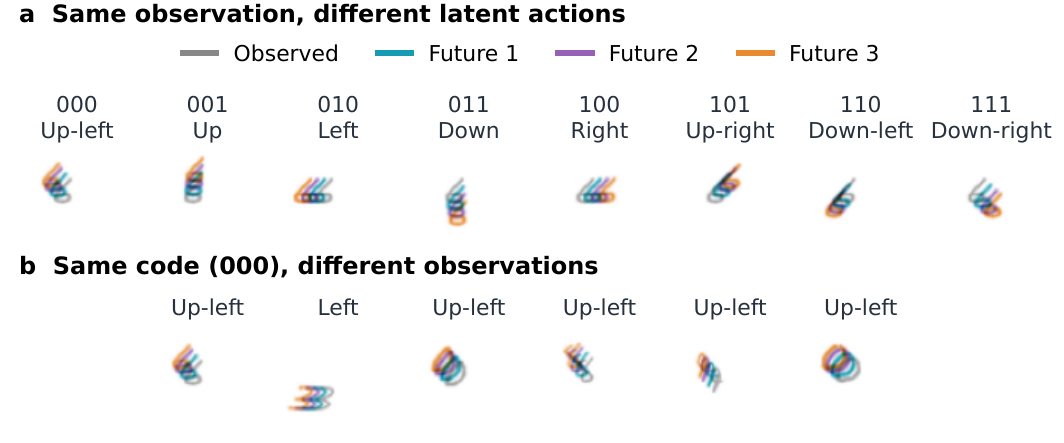}
\caption{Latent-action interventions using the ideal-gradient reference checkpoint.
(a) One observation, all eight codes. (b) Code 000, different digits.
Grey shows the observation; cyan, purple and orange show future frames 1--3.
Red labels mark invalid endpoints. Quantitative evaluation uses prior
sampling; direction frequencies appear in Appendix~\ref{app:metrics}
(\cref{fig:mnist-probabilities}).}
\label{fig:mnist-interventions}
\end{figure}

Both quantum circuit structures learn multiple image continuations under finite-shot parameter-shift training with readout noise (\cref{tab:mnist}). Their validity, trajectory errors and energy scores remain close to the strongest within-framework classical priors. Softmax-Categorical has better frequency matching, while QCBM uses fewer prior parameters (\cref{tab:parameters}). Product-QCBM performs comparably to Entangled-QCBM on this task: the target frequencies admit a product-bit representation, so entanglement is not required. The paired training comparisons in Appendix~\ref{app:gradient-comparison} show that these conclusions survive replacing ideal gradients with measurement-based estimates. \Cref{fig:mnist-interventions} shows the continuations produced by different latent-action codes, including unsuccessful predictions. Changing the code redirects the continuation, but its direction need not be identical across digit images.

Appendix~\ref{app:conditional-prior} tests whether the Born prior adapts
to different direction probabilities for different digits.

\subsection{Coordinated prediction on HEP18}
\begin{table}[tbp]
\centering\normalsize
\setlength{\tabcolsep}{2pt}
\medmuskip=0mu
\caption{HEP18 (128 observations, 128 draws each). Mean $\pm$ sample s.d. over three trainings; coverage/validity in percent, minADE/ES in metres. The first two rows are external methods. Training and highlighting follow Table~\ref{tab:mnist}.}
\label{tab:intersection}
\begin{tabular}{@{}lrrrrr@{}}
\toprule
Method / prior & \multicolumn{1}{c}{Cov. $\uparrow$} & \multicolumn{1}{c}{Valid $\uparrow$} & \multicolumn{1}{c}{TV $\downarrow$} & \multicolumn{1}{c}{minADE $\downarrow$} & \multicolumn{1}{c}{ES $\downarrow$} \\
\midrule
Variational JEPA & $0.00\,\pm\,0.00$ & $0.00\,\pm\,0.00$ & $1.000\,\pm\,0.000$ & $5.68\,\pm\,0.51$ & $44.78\,\pm\,1.59$ \\
Latent-CVAE & $59.82\,\pm\,4.73$ & $14.64\,\pm\,1.65$ & $0.854\,\pm\,0.016$ & $0.60\,\pm\,0.09$ & $34.54\,\pm\,0.07$ \\
\midrule % external-method separator
Deterministic JEPA & $2.68\,\pm\,2.78$ & $48.18\,\pm\,50.10$ & $0.973\,\pm\,0.028$ & $15.19\,\pm\,0.16$ & $64.58\,\pm\,0.68$ \\
Uniform-Bernoulli & $29.41\,\pm\,1.53$ & $6.53\,\pm\,0.15$ & $0.935\,\pm\,0.001$ & $1.40\,\pm\,0.07$ & $40.86\,\pm\,0.02$ \\
Softmax-Categorical & $\mathbf{99.97}\,\pm\,0.03$ & $\mathbf{98.00}\,\pm\,0.42$ & $\mathbf{0.157}\,\pm\,0.005$ & $\underline{0.20}\,\pm\,0.03$ & $\underline{34.34}\,\pm\,0.05$ \\
Independent-Bernoulli & $41.32\,\pm\,4.56$ & $25.66\,\pm\,2.49$ & $0.764\,\pm\,0.020$ & $4.37\,\pm\,0.67$ & $42.63\,\pm\,0.65$ \\
AR-GRU & $94.16\,\pm\,9.94$ & $95.54\,\pm\,0.85$ & $0.206\,\pm\,0.052$ & $0.28\,\pm\,0.15$ & $34.35\,\pm\,0.03$ \\
Mixture-4 & $\underline{99.96}\,\pm\,0.04$ & $\underline{97.88}\,\pm\,1.80$ & $\underline{0.159}\,\pm\,0.009$ & $\mathbf{0.19}\,\pm\,0.02$ & $\mathbf{34.30}\,\pm\,0.02$ \\
\midrule % prior-family separator
Frozen-QCBM & $29.24\,\pm\,1.75$ & $6.65\,\pm\,0.66$ & $0.933\,\pm\,0.007$ & $1.36\,\pm\,0.11$ & $40.79\,\pm\,0.12$ \\
\rowcolor{quantumrow}
Product-QCBM & $38.77\,\pm\,5.13$ & $24.50\,\pm\,4.11$ & $0.777\,\pm\,0.033$ & $4.23\,\pm\,0.49$ & $42.45\,\pm\,0.53$ \\
\rowcolor{quantumrow}
Entangled-QCBM & $99.73\,\pm\,0.25$ & $93.38\,\pm\,6.04$ & $0.188\,\pm\,0.036$ & $\underline{0.20}\,\pm\,0.02$ & $34.36\,\pm\,0.03$ \\
\bottomrule
\end{tabular}
\end{table}

Entangled-QCBM recovers nearly all coordinated futures under noisy parameter-shift training (\cref{tab:intersection}). It outperforms Product-QCBM and Frozen-QCBM, while the strongest classical priors retain higher mean validity.

For the three Entangled-QCBM runs in \cref{tab:intersection}, re-pairing agents within each test observation reduces mean joint validity from 93.38\% to 12.71\%. All four agents are individually valid in 96.61\% of original scenes, showing that local plausibility alone does not ensure coordination. We use ten independent permutations per observation (Appendix~\ref{app:metric-definitions}).

Trajectory errors and energy scores remain close to compact classical correlated priors; Appendix~\ref{app:gradient-comparison} separates finite-shot and readout-noise effects.

Qualitative examples are retained in Appendix~\ref{app:protocol} (\cref{fig:intersection}).

\subsection{Alternative routes from a common starting point}
\label{sec:d3il-results}
\begin{table}[tbp]
\centering\normalsize
\setlength{\tabcolsep}{3pt}
\medmuskip=0mu
\caption{D3IL from one direction-free starting point, evaluated with all three folds and three training seeds per method. Each checkpoint supplies 384 raw rollouts. We compute exact expected route statistics for a uniformly selected subset of 60 geometrically accepted paths. Entries are mean $\pm$ sample s.d. over three seed-level means, each averaging three folds; coverage is a percentage of 24 routes. Six deterministic checkpoints have no accepted paths, giving zero coverage and undefined TV. Quantum models use finite-shot parameter-shift training and 0.5\% readout flips. Metrics describe accepted-path diversity, not raw generation success.}
\label{tab:d3il}
\begin{tabular}{@{}lrrr@{}}
\toprule
Method / prior & \multicolumn{1}{c}{Routes $\uparrow$} & \multicolumn{1}{c}{Coverage $\uparrow$} & \multicolumn{1}{c}{TV $\downarrow$} \\
\midrule
Deterministic JEPA & $0.33\,\pm\,0.33$ & $1.39\,\pm\,1.39$ & -- \\
Latent-CVAE & $14.31\,\pm\,0.71$ & $59.63\,\pm\,2.96$ & $0.542\,\pm\,0.008$ \\
Softmax-Categorical & $18.08\,\pm\,0.89$ & $75.33\,\pm\,3.71$ & $0.401\,\pm\,0.028$ \\
AR-GRU & $17.65\,\pm\,0.69$ & $73.54\,\pm\,2.87$ & $0.412\,\pm\,0.018$ \\
Mixture-4 & $17.77\,\pm\,0.18$ & $74.05\,\pm\,0.75$ & $0.419\,\pm\,0.014$ \\
\midrule
Frozen-QCBM & $17.37\,\pm\,0.68$ & $72.36\,\pm\,2.82$ & $0.430\,\pm\,0.011$ \\
\rowcolor{quantumrow}
Product-QCBM & $18.07\,\pm\,0.59$ & $75.31\,\pm\,2.44$ & $0.408\,\pm\,0.022$ \\
\rowcolor{quantumrow}
Entangled-QCBM & $18.01\,\pm\,0.09$ & $75.06\,\pm\,0.37$ & $0.419\,\pm\,0.005$ \\
\bottomrule
\end{tabular}
\end{table}

\begin{figure}[!t]
\centering
\includegraphics[width=\textwidth]{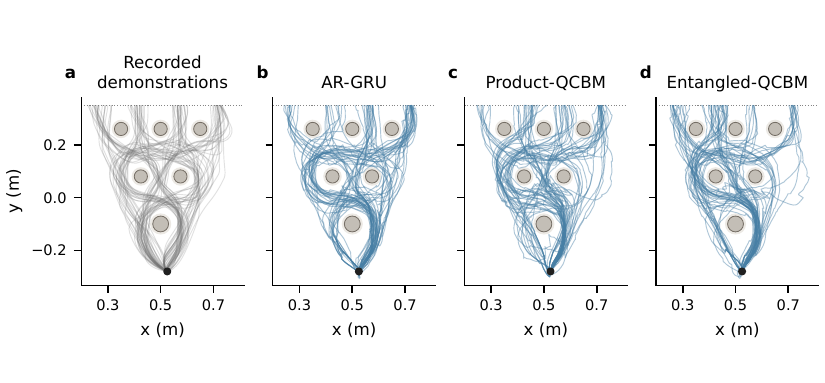}
\caption{D3IL continuations from one direction-free starting point
(black dot). The reference panel contains all 96 recorded demonstrations.
Each prediction panel shows 64 accepted paths from the first training
seed and first fold, selected from 384 draws without route balancing.
\Cref{tab:d3il} instead aggregates all nine models per method at a
common 60-accepted-path budget.
All panels use common axes; grey discs mark obstacles and pale rings
the scoring margin. Geometric acceptance does not ensure smooth motion,
and residual bends are retained. Appendix~\ref{app:d3il-qualitative}
shows all eight methods.}
\label{fig:d3il-trajectories}
\end{figure}

Our trained quantum priors generate multiple obstacle-crossing routes even
when the initial observation contains no motion direction
(\cref{tab:d3il,fig:d3il-trajectories}). Across the repeated models,
both trainable QCBMs recover a comparable number of routes to Softmax-Categorical
and AR-GRU. Softmax-Categorical has the lowest mean TV, with Product-QCBM
and AR-GRU close behind.
Each deterministic checkpoint returns only one repeated path, so it
cannot express multiple alternatives from this input.

Each trainable QCBM used 4,792 prior parameters versus Softmax-Categorical's 15,792,
a 69.7\% reduction. The complete deployed model was 4.2\% smaller,
since the encoder, predictor and readout dominated storage.
AR-GRU used 10,641 prior parameters, and Mixture-4 used 5,012.

\paragraph{Product-QCBM versus entangled sampling.}
Product-QCBM covers more routes in the illustrated 64-path sample,
but the repeated-model analysis gives similar coverage for the two
circuit families. Appendix~\ref{app:d3il-metrics} separates variation
across trained models from selection of a finite display sample.
More fundamentally, independent latent bits do not require independent
trajectory choices: the shared nonlinear predictor and recursive
re-encoding can map product-distributed codes into correlated routes.
This differs from HEP18, where separate agent predictors restrict how
stochastic coordination can arise. Product-QCBM sampling is consequently
a sufficient source of diverse latent actions for this D3IL setting;
the results do not establish a stable advantage from adding entanglement.

We use geometric acceptance only to define the route-distribution analysis,
not to alter generation or optimize the models. No collision penalty or
obstacle-based correction is used during training or recursive prediction.
The current protocol measures route diversity conditional on geometric
acceptance; it does not evaluate a learned collision-avoidance objective.
Its frequency estimates concern one initial input and accepted trajectories.
Training with the same stationary first input
would remove the initial-input mismatch tested here.

\subsection{Comparison of complete learning methods}
Our correlated classical and quantum implementations improve validity,
TV and energy score over both external adaptations
(\cref{tab:mnist,tab:intersection}), identifying a framework-level benefit.
Latent-CVAE has higher coverage and lower minimum ADE on MovingMNIST
(\cref{tab:mnist}),
but many draws fall between valid continuations. On HEP18, our correlated
models also improve coverage and trajectory error.
Appendix~\ref{app:external-analysis} discusses these benchmark adaptations
and their distance--coordination trade-offs.

\subsection{Real-QPU inference}
\label{sec:qpu}
We ran trained quantum priors on real QPUs and used their sampled actions
in the complete prediction pipeline. Hardware noise can change the
latent-action distribution and, in turn, the frequencies of predicted
futures. This shift can improve or reduce individual benchmark metrics;
its effect depends on the learned distribution, task and metric.
On MovingMNIST, the reference
RY/CNOT and Product-QCBM circuits
closely matched the ideal-sampling results (\cref{tab:qpu}).

For HEP18's 12-qubit circuit, hardware errors reduced joint prediction
validity. We improved execution by choosing better-performing physical
qubits and connections, without retraining or changing the circuit's ideal
distribution. Joint validity rose from 76.88\% to 86.69\%, compared with
98.41\% under ideal sampling. After optimization, coverage and trajectory
errors were close to the ideal reference (\cref{tab:hep18-qpu-optimized}).
Appendix~\ref{app:qpu} details these hardware implementations and
their matched evaluations.

On D3IL, the trained Product-QCBM and Entangled-QCBM priors generated
the latent actions for complete recursive trajectory prediction.
Their route coverage fell within the ranges observed across matched
GPU sampling repetitions.
Each recursive step requires a new conditional circuit execution, so we
used a smaller hardware budget of 40 raw rollouts per prior.
We matched the GPU runs to this budget; the main D3IL comparison
uses 384 raw rollouts per checkpoint.
Appendix~\ref{app:d3il-qpu} explains the resulting differences in
sample counts and reports the matched comparisons.

\section{Conclusion}
We introduce \method{}, which trains a conditional Born prior through the
predictive consequences of sampled latent actions. Our model learns multimodal
and coordinated futures on constructed tasks and produces recursive
predictions from diverse human demonstrations in D3IL under finite-shot,
noisy training. On the controlled tasks, our framework improves over the
evaluated variational baselines through classical and quantum prior implementations.
On D3IL, the trained quantum priors produce diverse routes from a common
direction-free starting point, with route coverage comparable to classical
discrete priors in this fixed-input diagnostic. Within this framework,
our quantum priors offer competitive prediction quality and a
parameter-efficient representation of binary action distributions.
We validate finite-shot parameter-shift training and real-QPU inference,
establishing a practical route for quantum generative models to participate
in representation-space future prediction.

\paragraph{Limitations and future work.}
Our experiments establish training and inference feasibility, but the quantum
priors do not consistently outperform strong classical methods across all
predictive metrics. Our current circuits are shallow, with performance
and optimization stability depending on the task and circuit.
Larger implementations face hardware noise and substantial measurement costs.
We will develop circuit structures and training
algorithms tailored to multimodal prediction and correlated latent actions,
and investigate quantum implementations of more of the JEPA predictor.
We will also formulate latent-action reasoning and planning as optimization
problems for special-purpose quantum platforms.

\section*{Reproducibility Statement}
We specify the objective and prior update in \cref{sec:method}; data
construction, observation-level splits, readouts and metrics in
Appendix~\ref{app:protocol}; architectures, circuits, optimization and
parameter counts in Appendices~\ref{app:training} and \ref{app:circuits};
evaluation sensitivity and direction frequencies in Appendix~\ref{app:metrics};
external-method adaptations in Appendix~\ref{app:external}; and QPU procedures
in Appendices~\ref{app:qpu} and \ref{app:d3il-qpu}.
Appendix~\ref{app:d3il-implementation} describes the public-demonstration
study, including its training stages, data splits and recursive evaluation.

\bibliographystyle{iclr2027_conference}
\bibliography{references}
\clearpage
\appendix
% Appendix-only float policy: share pages with text and avoid centred,
% partially empty float pages. Main-text geometry and spacing are unchanged.
\setcounter{topnumber}{3}
\setcounter{bottomnumber}{2}
\setcounter{totalnumber}{5}
\renewcommand{\topfraction}{0.95}
\renewcommand{\bottomfraction}{0.85}
\renewcommand{\textfraction}{0.05}
\renewcommand{\floatpagefraction}{0.85}
\makeatletter
\setlength{\@fptop}{0pt}
\setlength{\@fpsep}{12pt}
\setlength{\@fpbot}{0pt plus 1fil}
\makeatother
\raggedbottom
\section*{Appendix}
\addcontentsline{toc}{section}{Appendix}

\section{Implementation and Training Details}
\label{app:training}
\subsection{Classical prior architectures}
All learned classical priors receive the observation representation $h$.
Here $d$ is its dimension, $H_p$ is the prior's hidden width, and $n$
is the number of action bits.

\paragraph{Softmax-Categorical.}
We apply layer normalization, a $d\!\rightarrow\!H_p$ linear layer,
SiLU, and an $H_p\!\rightarrow\!2^n$ output layer.
The output gives one logit for every possible binary code.
We draw a categorical index and convert it to an $n$-bit string.
We evaluate this prior at both three and twelve bits.
At twelve bits, its explicit enumeration of code probabilities requires
more parameters than the other priors.

\paragraph{Independent-Bernoulli.}
This prior uses the same conditioner architecture, but its output layer
produces $n$ logits.
Each logit defines one bit's probability of being one, and we sample
the bits independently.

\paragraph{AR-GRU.}
We project the layer-normalized observation representation into a
width-$H_p$ hidden state.
A shared GRU cell generates the bits sequentially, receiving the
previous bit as input. The initial input bit is zero, and a shared
scalar output head predicts the next Bernoulli logit.
Later bits can therefore depend on both the observation and earlier
sampled bits.

\paragraph{Bernoulli mixture (Mixture-$R$).}
A SiLU MLP maps the layer-normalized observation through widths
$d\!\rightarrow\!H_p\!\rightarrow\!R(n+1)$.
Its outputs are $R$ component logits and $Rn$ Bernoulli logits.
We first sample one mixture component, then sample all bits independently
using that component's Bernoulli probabilities.
For the training update, we compute each sampled code's probability
by summing over components, using log-sum-exp for numerical stability.

The main HEP18 comparison uses $R=4$, with 5,796 allocated prior
parameters, compared with 5,628 for the QCBM.
We also report $R=2$ and $R=8$ to show how predictive accuracy
changes with mixture capacity.
MovingMNIST uses $R=2$, with 3,640 prior parameters, close to the
three-bit QCBM's 3,435.
Here $R$ counts mixture components, not future modes.
Each component can assign probability to every one of the $2^n$
binary codes.

\subsection{Benchmark encoders, predictors and readouts}
\label{app:networks}
\paragraph{MovingMNIST.}
The shared encoder takes a three-frame clip, stacking the frames as
input channels. It then applies
convolutions $3\!\rightarrow\!24$ ($5\times5$),
$24\!\rightarrow\!48$ ($3\times3$), and
$48\!\rightarrow\!64$ ($3\times3$).
All three layers use stride 2 and SiLU, with padding of 2, 1 and
1 pixels, respectively.
These layers reduce the spatial resolution from $64\times64$ to $8\times8$.
We flatten the resulting $64\times8\times8$ feature grid and map it
through fully connected widths 96 and 64, followed by layer normalization.
The predictor follows \cref{eq:code,eq:predictor} with width 96 and
layer normalization at its output.
Appendix~\ref{app:circuits} describes the two-layer, three-qubit circuit.

The image readout maps $[h;t]$ through widths
$128\!\rightarrow\!96\!\rightarrow\!6$ with SiLU.
We interpret the six outputs as three two-dimensional displacements,
bounding each component by $18\tanh(a/18)$, where $a$ is the raw output.
At inference, the readout receives $\widehat t$ in place of $t$.
A bilinear spatial transformer applies each displacement to the last
observation image, with zero padding and aligned corners.

The readout minimizes weighted image L1 error,
$\operatorname{mean}[(1+4y)|\widehat y-y|]$, plus one half of the
soft Dice loss.
Here $y$ and $\widehat y$ are the target and predicted pixel intensities.
The L1 weight emphasizes digit pixels, while Dice measures image overlap.
We average over frames and use Dice smoothing $10^{-6}$.
The readout's final layer is initialized to zero, so its initial
predictions have zero displacement.
We train it for 4,000 iterations with batch size 64.
AdamW uses learning rate $0.002$, weight decay $0.01$, and gradient
clipping at norm 5.

\paragraph{HEP18.}
Each agent's encoder receives its three position--speed pairs as six scalars.
The shared MLP has hidden widths 64 and 64, SiLU activations,
and a 16-dimensional output.
Concatenating the four agent outputs gives the scene representation with $d=64$.
The residual predictor uses a width-64 local-code MLP.
Its output remains in the encoder's coordinates, without output normalization.

The trajectory readout is a shared $16\!\rightarrow\!64\!\rightarrow\!3$
SiLU MLP. It receives an agent's encoded target during readout training
and its predicted representation at inference.
The three outputs give speeds bounded by
$12\operatorname{sigmoid}(a)$\,m/s, where $a$ denotes the raw outputs.
We recover positions using the fixed integration procedure in
Appendix~\ref{app:intersection}.
The readout loss sums position and speed MSE, after scaling position
errors by 20\,m and speed errors by 5\,m/s.
We train for 16,000 AdamW steps with batch size 128, learning rate
0.002 and weight decay 0.01.

\subsection{Optimization and gradient estimation}
\begin{table}[tbp]
\centering\small
\setlength{\tabcolsep}{3pt}
\caption{Reference training configuration for
\cref{tab:mnist,tab:intersection}. Readout training is a separate
stage with all JEPA parameters frozen. These trained QCBMs use
finite-shot parameter shift and a 0.5\% independent readout-flip
probability per bit.}
\label{tab:configuration}
\begin{tabular}{@{}lrr@{}}
\toprule
Setting & \multicolumn{1}{c}{MovingMNIST} & \multicolumn{1}{c}{HEP18} \\
\midrule
Latent dimension $d$ / predictor width & 64 / 96 & 64 / 64 \\
Prior width $H_p$ & 48 & 48 \\
Bits $n$ / body layers $L$ & 3 / 2 & 12 / 4 \\
Terminal $R_Y$ angles & 3 & 12 \\
World-model iterations & 4,000 & 6,000 \\
Batch size / prior samples & 64 / 32 & 32 / 32 \\
Target views per observation in a batch & 1 & 18 \\
Representation-only / fixed-prior warm-up & 200 / 300 & 200 / 300 \\
Encoder learning rate, initial / later & $3\,10^{-4}$ / $10^{-5}$ &
 $3\,10^{-4}$ / $10^{-5}$ \\
Predictor / prior learning rate & $10^{-3}$ / $10^{-3}$ &
 $10^{-3}$ / $10^{-3}$ \\
Readout iterations / batch & 4,000 / 64 & 16,000 / 128 \\
Readout learning rate & $2\,10^{-3}$ & $2\,10^{-3}$ \\
Shifted observations $b$ / batch $B$ & 8 / 64 & 8 / 32 \\
Shifted coordinates $r$ / total $G$ & 9 / 9 & 12 / 60 \\
Shots per shifted circuit & 256 & 512 \\
\bottomrule
\end{tabular}
\end{table}

We use AdamW with weight decay $10^{-5}$ for world-model training.
Before each update, we clip each parameter group's gradient norm at 5.
We set $\lambda_{\rm reg}=4/3$, which gives $R(h)$ and $R(t)$
a coefficient of $2/3$ each.
The encoder learning rate decreases after 200 iterations, when predictive
training begins. Prior updates start after 300 iterations.
\Cref{tab:configuration} lists the learning rates, iteration counts and sampling budgets.

\paragraph{VISReg implementation.}
\label{app:visreg}
For a feature batch $v\in\mathbb R^{B\times d}$, each row is one
observation and each column is one feature coordinate.
We compute the coordinate-wise mean $\mu$ and population standard deviation
$s$ across the $B$ rows.
For the shape term, we normalize features as
$\widetilde v=(v-\mu)/(\sg(s)+\epsilon)$, with $\epsilon=10^{-6}$.
The stop-gradient operator $\sg$ treats $s$ as fixed in this normalization.
We also draw $J$ independent Gaussian vectors and normalize them to
unit directions $w_j$.
The three components of the regularizer in \cref{eq:regularizer} are
\begin{align*}
 \mathcal L_{\rm center}(v)&=\|\mu\|_2^2/d,\qquad
 \mathcal L_{\rm scale}(v)=\|s-\mathbf1\|_2^2/d,\\
 \mathcal L_{\rm shape}(v)&=\frac1{BJ}\sum_{j=1}^{J}\sum_{i=1}^{B}
 \left([\operatorname{sort}(\widetilde v w_j)]_i
       -\Phi^{-1}\!\left(\frac{i}{B+1}\right)\right)^2,
\end{align*}
Here $\Phi$ is the standard-normal cumulative distribution function,
and $\Phi^{-1}$ gives its quantiles.
The centre term moves each coordinate's batch mean towards zero.
The scale term moves each coordinate's standard deviation towards one.
For the shape term, we project the normalized features onto each
direction $w_j$ and sort the resulting $B$ values.
We compare these ordered projections with Gaussian quantiles at probabilities
$i/(B+1)$, then average their squared differences.

Our implementation averages over coordinates and projected samples.
The regularizer weights therefore absorb normalization factors that appear
in norm-based expressions of VISReg.
We use $J=128$, $\lambda_{\rm center}=1$ and $\lambda_{\rm scale}=10$;
$\lambda_{\rm shape}$ increases linearly from zero to one over the first
100 iterations.
All priors use these same weights and schedule, with the original
VISReg component functions.
Only the standard deviation in the shape normalization is detached;
the scale term still differentiates through $s$.
Both observation and target encoder branches receive gradients from
representation regularization.

\subsection{Classical and ideal-gradient reference updates}
\label{app:gradient}
Classical priors use the leave-one-out score-function update in
\cref{eq:prior-surrogate}.
For each sampled code, this update combines its predictive cost with
the gradient of its log probability.
When an observation has several target futures, we average the
prediction--target distance over those futures.
Each classical prior computes the code log probability using its
own parameterization.

The ideal-gradient QCBM reference uses the same sampled-cost update,
but differentiates Born log probabilities through the simulator.
Gradients reach the classical conditioner, body-angle offsets and terminal
rotation angles.
These references still sample codes for the predictive objective;
they do not enumerate predictions for every possible code.
The trainable QCBMs in \cref{tab:mnist,tab:intersection,tab:d3il} replace this
simulator gradient with the measurement-based update in
\cref{sec:prior-training}. Ideal-gradient circuit comparisons and earlier
D3IL ablations are identified separately in their respective appendices.

\paragraph{Prior-gradient estimator.}
The prior affects both the target-fit term and the pairwise diversity
term in the energy score.
To see the resulting derivative, write a discrete objective as
$a^\top p+p^\top Bp$, with probability vector $p$ and symmetric matrix $B$.
Here $a$ contains prediction--target distances, while $B$ contains the
negative half-distances between pairs of predictions.
The derivative with respect to $p$ is $a+2Bp$, giving
\cref{eq:firstvariation} after substituting the energy terms.
This explains why the prior's cost uses the full pairwise-distance
expectation, even though the energy loss contains a factor of one half.
This vector notation explains the derivative; our sampled update does
not construct these probability or distance tables.

The leave-one-out estimate in \cref{eq:prior-surrogate} compares each
sample with the other $K-1$ predictions, excluding its self-pair.
For the prior update, we detach this cost and differentiate only
the sampled code's log probability.
Classical priors and ideal-gradient QCBM references use this estimator.
The reference QCBMs in \cref{tab:mnist,tab:intersection} instead use
shifted-circuit measurements to estimate the
same prior derivative through \cref{eq:shift}.

During parameter shift, we hold the reference predictions fixed and
evaluate the cost under positive and negative angle shifts.
Independent reference and shifted samples give the population derivative
in expectation.
The pairwise term depends on two draws from the prior.
Holding one reference distribution fixed lets the shifted circuits
estimate the derivative through the other draw.
The factor of two in the cost accounts for both draws'
dependence on the prior.
For the RY/CNOT reference, we uniformly select $r$ of the
$G=n(L+1)$ rotation coordinates and
$b$ of the $B$ observations in a batch.
The factors $G/r$ and $B/b$ correct for the coordinates and observations
that were not evaluated.
\Cref{tab:configuration} gives the subsampling and shot budgets used for each task.

\paragraph{Training sequence.}
Once the warm-up stages are complete, each world-model iteration proceeds
as follows:
\begin{enumerate}
\item We draw a batch of observations and their realized future views,
then encode both with the shared encoder.
\item We sample $K$ actions per observation and predict one future
representation for each sampled action.
\item We compute predictive energy with the target representations
detached, and representation regularization with both encoder branches trainable.
For HEP18, we also regularize the change representations as specified
in Appendix~\ref{app:intersection}.
\item We compute a separate prior gradient, using parameter shift for
QCBMs or the score-function estimator for classical priors.
This update changes the prior only; it holds the encoded observations,
targets and predictor fixed.
\item We combine the gradients for each parameter group, clip its
norm at 5, and take an AdamW step.
\end{enumerate}
Iterations 1--200 train the encoder using only representation regularization.
Iterations 201--300 also train the predictor, while retaining the
initialized prior.
Later iterations update all trainable groups; frozen-prior controls
continue to omit prior updates.
After world-model training, we freeze all JEPA parameters for readout
training and evaluation.
We use the final checkpoints, without selecting them by test performance.

\section{D3IL Data, Models and Offline Evaluation}
\label{app:d3il-implementation}
\subsection{Demonstrations and input construction}
We use 96 public human demonstrations of the D3IL Avoiding task,
recorded in a simulated obstacle course \cite{jia2024d3il}.
Each of three route-stratified folds assigns 64 complete demonstrations
to training and 32 to evaluation. Route labels are used only to
construct these splits and evaluate predictions.

We construct the first training observation by interpolating recorded
frames 4--8 to 24 points. For subsequent predictions, the observation
is the preceding segment.
We divide futures into 8\,cm arc-length segments, each resampled to
24 points, retaining shorter final segments and excluding padding.
This gives 1,030 segments from the 96 source demonstrations.
Coordinates are normalized as $(\mathrm{xy}-(0.5,-0.28))/0.4$;
evaluation restores metres. Training and evaluation use separate
demonstrations within each fold.

\subsection{Encoder, predictor and readout}
The shared encoder forms seven-dimensional tokens from position, first
and second differences, and a normalized sequence index.
A linear layer, layer normalization and SiLU map tokens to width 128,
followed by a width-128 GRU. Its final and mean states are concatenated
and projected to 64 dimensions.
A parallel $16\!\rightarrow\!128\!\rightarrow\!64$ SiLU MLP encodes
kinematic summaries: final displacement; displacement mean and standard
deviation; first-difference mean, standard deviation and final value;
and second-difference mean and standard deviation.
Each quantity has two coordinates. We add this branch to the GRU output.

The width-128 predictor uses the shared eight-bit code embedding in
\cref{eq:code,eq:predictor}. D3IL additionally uses a two-output head
from its hidden activation, setting $s=2\operatorname{sigmoid}(a)$ and
$b=0.25\tanh(c)$. For layer-normalized residual output $q$ with affine
parameters $w_{\rm LN},b_{\rm LN}$, its prediction is
\begin{equation}
 \widehat t=q+(s-1)(q-b_{\rm LN})+w_{\rm LN}b.
\end{equation}
Zero initialization gives $s=1,b=0$. This bounded scale and offset head
is shared by all discrete-prior and deterministic D3IL comparisons;
MovingMNIST and HEP18 do not use it.

The readout is a $128\!\rightarrow\!128\!\rightarrow\!128\!\rightarrow\!16$
SiLU MLP receiving $[h;t]$, or $[h;\widehat t]$ at inference.
Its outputs are eight two-dimensional control points for a clamped
cubic B-spline. We evaluate the spline at 24 evenly spaced parameter
values, excluding the starting point and including the endpoint.
The readout imposes no obstacle or path-length constraints.
\Cref{tab:d3il-parameters} gives model sizes.

\subsection{Training and comparison methods}
The sampled models use normalized Euclidean energy with regularization
$(2/3)[R(h)+R(t)]$.
The initial JEPA stage follows the AdamW, warm-up, learning-rate and
VISReg settings in Appendix~\ref{app:training}, for 12,000 updates with
batch size 64 and 32 prior samples.
For each quantum prior update, we select eight observations and twelve
of forty rotation angles uniformly without replacement.
Each signed shift uses 512 shots plus 32 independent unshifted reference
draws per selected observation. We scale coordinate gradients by $40/12$
and average over the selected observations.
The noise model applies independent 0.5\% readout flips to forward,
shifted and evaluation samples.

Next, we freeze the JEPA and train the readout on encoded real targets
with Smooth L1 coordinate loss. We use 8,000 steps at learning rate
0.002 and 2,000 at 0.0002. Batch size is 64, AdamW weight decay is $10^{-5}$, and gradient
norms are clipped at five.
Finally, we freeze encoder and readout and train predictor and prior
for 1,000 steps at rate $10^{-4}$. Sixteen real three-segment windows
provide 48 local observation--future pairs per batch.
The same energy and prior updates apply; quantum gradients select eight
pairs under the same measurement settings.
This last stage trains on recorded segment pairs rather than generated
rollouts. The training objective contains no obstacle penalties.

Frozen-QCBM fixes the entire initialized prior throughout both predictive
stages. Its zero-initialized conditioner output remains zero, so its
distribution is input-independent; the predictor still receives the observation.
The trainable QCBMs have 11,700 initial and 1,000 final prior updates,
consuming 1,277,824,000 simulated training shots per run.
Frozen-QCBM uses 26,112,000 forward shots without shifts.
These shot counts cover training in the simulator and exclude evaluation.

\paragraph{Prior implementations.}
Entangled-QCBM has four RY/CNOT body layers and terminal RY rotations:
forty rotations and fourteen CNOTs.
Even-indexed layers couple $(0,1),(2,3),\ldots$;
odd-indexed layers couple $(1,2),(3,4),\ldots$, with lower-index controls.
The conditioner has widths $64\!\rightarrow\!48\!\rightarrow\!32$.
Product-QCBM removes CNOTs; Frozen-QCBM fixes the entangled initialization.
AR-GRU uses the common width-48 prior in Appendix~\ref{app:training};
Mixture-4 has four Bernoulli components, and Softmax-Categorical has 256 logits.
All six sampled priors share the encoder, predictor, readout and schedule.
Deterministic JEPA fixes the code to zero and uses representation MSE;
it retains the final predictor-only training stage.

\paragraph{Latent-CVAE.}
The twelve-dimensional Gaussian prior has widths
$64\!\rightarrow\!48\!\rightarrow\!24$, and the future-conditioned
posterior has widths $128\!\rightarrow\!128\!\rightarrow\!24$.
A residual $76\!\rightarrow\!128\!\rightarrow\!64$ transition predicts features.
Its objective combines target-feature MSE, dimension-averaged Gaussian
KL with weight one, and the same two-view VISReg.
It uses 13,000 Adam updates at a fixed learning rate of $10^{-3}$,
followed by the common 10,000-step readout fit.
The posterior's 19,608 parameters are used only during training.

We evaluate the final checkpoint of every training run.
The resource table distinguishes prior size from full deployed size.
Each trainable QCBM allocates 4,792 prior values, including 64 inactive
coordinates; Frozen-QCBM retains these values at deployment despite
having no trainable prior parameters.
\begin{table}[tbp]
\centering\small
\setlength{\tabcolsep}{3pt}
\medmuskip=0mu
\caption{D3IL model resources. The first three numeric columns count trainable parameters in their respective stages; deployment includes fixed prior weights but excludes the CVAE posterior. Frozen-QCBM stores 4,792 fixed prior values in addition to its trainable neural modules. Quantum counts include 64 inactive coordinates retained for checkpoint compatibility. AR-GRU uses the same width-48 recurrent prior as MovingMNIST and HEP18.}
\label{tab:d3il-parameters}
\begin{tabular}{@{}lrrrr@{}}
\toprule
Method / prior & \multicolumn{1}{c}{Prior} & \multicolumn{1}{c}{JEPA training} & \multicolumn{1}{c}{Readout} & \multicolumn{1}{c}{Deployment} \\
\midrule
Deterministic JEPA & 0 & 211,266 & 35,088 & 246,354 \\
Latent-CVAE & 4,296 & 169,248 & 35,088 & 184,728 \\
Softmax-Categorical & 15,792 & 227,058 & 35,088 & 262,146 \\
AR-GRU & 10,641 & 221,907 & 35,088 & 256,995 \\
Mixture-4 & 5,012 & 216,278 & 35,088 & 251,366 \\
Frozen-QCBM & 0 & 211,266 & 35,088 & 251,146 \\
\rowcolor{quantumrow}
Product-QCBM & 4,792 & 216,058 & 35,088 & 251,146 \\
\rowcolor{quantumrow}
Entangled-QCBM & 4,792 & 216,058 & 35,088 & 251,146 \\
\bottomrule
\end{tabular}
\end{table}

\subsection{Common-start generation and route coverage}
\label{app:d3il-metrics}
All 96 raw trajectories have exactly the same starting point,
approximately $(0.52492847,-0.27973673)$\,m.
For evaluation, we repeat that point 24 times, giving no observed motion
or heading. Sequence indices supply token order only.
We encode this observation, sample a code, predict and decode a segment,
then re-encode its coordinates before the next call.
Generation stops when the trajectory first crosses $y=0.35$\,m or
after sixteen calls. We interpolate the endpoint at the crossing.
Every trajectory starts directly from the stationary point.
This differs from the moving observations used for training.
We evaluate the resulting trajectories offline.

A path is accepted only if it reaches the goal, crosses every obstacle
row, avoids all inflated obstacles and stays within the geometric reach
envelope. We include starting and inter-segment connections and the final
partial edge in continuous collision checks.
Obstacle radii add 1\,cm for the manipulator tip; the reach bound is
approximately 1.0913\,m with tolerance $10^{-6}$\,m.
Acceptance depends on this planar geometry; robot dynamics, orientation
and joint limits are not evaluated.

We classify routes by the first upward crossing of each obstacle-row
centre line. The rows have two, three and four gaps, yielding
$2\times3\times4=24$ routes. For a set of accepted paths, let $N_r$
count route $r$, $N=\sum_rN_r$, and $C$ be the number of routes
represented by at least one path. Then
\begin{equation}
 \mathrm{Coverage}=\frac{C}{24},\qquad
 \mathrm{TV}=\frac12\sum_{r=1}^{24}
 \left|\frac{N_r}{N}-\frac1{24}\right|.
 \label{eq:d3il-route-metrics}
\end{equation}
Coverage is expressed as a percentage in the tables.
An empty accepted set has zero coverage and undefined TV.
The uniform reference approximates the recorded route frequencies:
22 routes have four demonstrations, one has three and one has five.
It includes training trajectories and is not a separately measured
conditional ground truth for the stationary input.

Each of the nine models per method supplies three 128-path repetitions,
forming a 384-raw-path pool. All stochastic pools have at least 61
accepted paths; one Latent-CVAE pool has fewer than 64.
We therefore use a common 60-accepted-path budget, without replenishing
pools or discarding checkpoints.
We compute expected coverage and TV over uniformly drawn 60-path subsets.
The route counts follow a hypergeometric distribution, allowing exact
expectations without repeated subsampling.
No route balancing or nearest-target selection is applied.
Six deterministic models have no accepted paths; the other three repeat
one accepted route. We therefore report the aggregate deterministic TV
as undefined.

We average folds within each seed, then report the mean and sample
standard deviation of the three seed-level means in \cref{tab:d3il}.
The three seed-level means measure training variation, while every
model is evaluated from the same starting point.
The metrics describe route diversity among geometrically accepted paths,
rather than the success rate of all generated paths.

\subsection{Qualitative comparison}
\label{app:d3il-qualitative}
The figure shows the first 64 accepted paths under one common random
ordering for the first fold and training run.
This illustrates one trained model per method; \cref{tab:d3il} summarizes
all nine models using the common 60-path budget.
All methods receive the same stationary point, and the reference contains
all 96 demonstrations. We retain bends and backtracking without smoothing.
\Cref{fig:d3il-trajectories} enlarges three of these method panels.

\begin{figure}[!htbp]
\centering
\includegraphics[width=\textwidth]{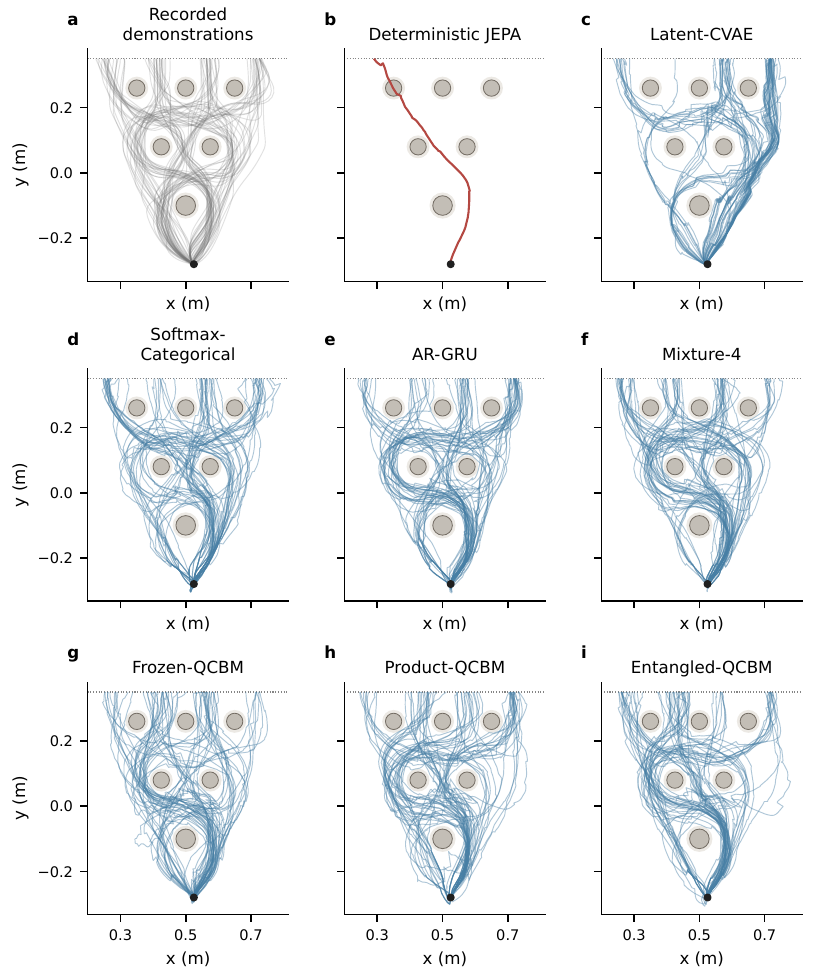}
\caption{D3IL from one common point, with no observed direction.
The reference shows all demonstrations; stochastic panels show 64
accepted paths per method. The single deterministic prediction is red
because it collides. All panels share the starting point (black dot),
axes and obstacle geometry. Paths are not balanced across routes or
smoothed. Trainable QCBMs use noisy parameter-shift training and 0.5\%
readout flips at evaluation.}
\label{fig:d3il-fold0}
\end{figure}

\section{Circuit Layouts and Parameter Accounting}
\label{app:circuits}
The MovingMNIST circuit has two body layers, followed by three terminal
$R_Y$ rotations.
The first body layer uses CNOT $(0,1)$, and the second uses CNOT $(1,2)$.
In each ordered pair, the first index is the control and the
second is the target. The complete circuit has nine single-qubit
rotations and two CNOTs.

For HEP18, we arrange the twelve qubits as four rows of three bits,
ordered north, west, south and east.
Row and column indices both start at zero.
The four body layers use the following CNOT connections:
\begin{enumerate}
\item Within every row, layer 1 couples columns $(0,1)$.
\item At every column, layer 2 couples rows $(0,1)$ and $(2,3)$.
\item Within every row, layer 3 couples columns $(1,2)$.
\item At every column, layer 4 couples rows $(1,2)$ and $(3,0)$.
\end{enumerate}
Each ordered pair again specifies control followed by target.
Twelve terminal rotations follow the fourth layer, giving a total of
60 $R_Y$ rotations and 20 CNOTs.

The Product-QCBM configuration removes all CNOT gates and retains the same
rotation parameters and conditioner.
All rotations on a given qubit then act about the same axis.
Only their summed angle affects that bit's Bernoulli probability.
The extra rotation layers preserve the matched parameterization, but
do not introduce correlations between bits.

For memory accounting, we distinguish allocated parameters from parameters
used in the forward computation.
The implementation allocates three body-angle coordinates per qubit
and layer, but this ansatz uses only the $Y$ coordinate.
HEP18 therefore has 96 unused body-angle entries, and MovingMNIST has 12.
All terminal angles are used in the forward computation.
The shared code predictor also includes an agent-identity offset,
including in the single-digit model.
\begin{table}[tbp]
\centering\small
\setlength{\tabcolsep}{5pt}
\caption{Parameter counts for MovingMNIST (Digits, 3-bit) and HEP18
(Traffic, 12-bit). The Prior column counts all allocated prior
parameters; Active counts only those used in the forward computation.
Frozen-QCBM has the same architecture as Entangled-QCBM but receives
no prior updates. Product-QCBM has the same active rotation count, although
only their sum on each qubit affects its output probability.}
\label{tab:parameters}
\begin{tabular}{@{}lrrrrr@{}}
\toprule
Task / prior & \multicolumn{1}{c}{Encoder} & \multicolumn{1}{c}{Predictor} & \multicolumn{1}{c}{Prior} & \multicolumn{1}{c}{Active} & \multicolumn{1}{c}{Readout} \\
\midrule
Digits / Deterministic JEPA & 439,600 & 50,112 & 0 & 0 & 12,966 \\
Digits / Uniform-Bernoulli & 439,600 & 50,112 & 0 & 0 & 12,966 \\
Digits / Frozen-QCBM & 439,600 & 50,112 & 3,435 & 3,423 & 12,966 \\
Digits / Softmax-Categorical & 439,600 & 50,112 & 3,640 & 3,640 & 12,966 \\
Digits / AR-GRU & 439,600 & 50,112 & 10,641 & 10,641 & 12,966 \\
Digits / Mixture-2 & 439,600 & 50,112 & 3,640 & 3,640 & 12,966 \\
Digits / Product-QCBM & 439,600 & 50,112 & 3,435 & 3,423 & 12,966 \\
Digits / Entangled-QCBM & 439,600 & 50,112 & 3,435 & 3,423 & 12,966 \\
\midrule
Traffic / Deterministic JEPA & 5,648 & 22,800 & 0 & 0 & 1,283 \\
Traffic / Uniform-Bernoulli & 5,648 & 22,800 & 0 & 0 & 1,283 \\
Traffic / Frozen-QCBM & 5,648 & 22,800 & 5,628 & 5,532 & 1,283 \\
Traffic / Softmax-Categorical & 5,648 & 22,800 & 203,952 & 203,952 & 1,283 \\
Traffic / Independent-Bernoulli & 5,648 & 22,800 & 3,836 & 3,836 & 1,283 \\
Traffic / AR-GRU & 5,648 & 22,800 & 10,641 & 10,641 & 1,283 \\
Traffic / Mixture-4 & 5,648 & 22,800 & 5,796 & 5,796 & 1,283 \\
Traffic / Product-QCBM & 5,648 & 22,800 & 5,628 & 5,532 & 1,283 \\
Traffic / Entangled-QCBM & 5,648 & 22,800 & 5,628 & 5,532 & 1,283 \\
Traffic / Mixture-2 & 5,648 & 22,800 & 4,522 & 4,522 & 1,283 \\
Traffic / Mixture-8 & 5,648 & 22,800 & 8,344 & 8,344 & 1,283 \\
\bottomrule
\end{tabular}
\end{table}

\subsection{Resource-scaling details}
\label{app:resource-details}
At fixed numerical precision, weight storage follows the parameter
counts in \cref{tab:resources}. Working memory excludes these weights
and the common $O(Kn)$ storage for sampled bit strings.
Softmax-Categorical, Independent-Bernoulli and Mixture-$R$ require
$O(2^n+H_p)$, $O(n+H_p)$ and $O(Rn+H_p)$ temporary storage,
respectively. AR-GRU requires $O(KH_p)$ recurrent-state storage when
generating $K$ codes in parallel, with $n$ sequential decisions per code.
A hardware QCBM uses $n$ qubits and $O(nL+H_p)$ classical conditioner
memory. Parallel two-qubit matchings give depth $O(L)$ per draw,
assuming compatible hardware connectivity.

\section{Finite-Shot Training and Readout-Noise Comparison}
\label{app:gradient-comparison}
We compare three gradient-estimation settings for each QCBM family:
ideal-gradient simulation, noise-free finite-shot parameter shift, and
parameter shift with independent 0.5\% readout flips per bit.
The noisy setting supplies the Product-QCBM and Entangled-QCBM rows in
the main tables, \cref{tab:mnist,tab:intersection}.
Each setting uses three independent initializations, matched across
settings, and we retain every training run.
We keep the architecture, data, losses, iteration counts, readout training
and 128-draw evaluation budget fixed.

The two parameter-shift settings estimate gradients from measured bit
strings rather than from differentiable simulated probabilities.
In the noisy setting, the same readout-flip channel is active during
both training and evaluation.
Its results therefore reflect noise in both gradient estimation and
prediction-time sampling.

For each selected observation and angle, we sample circuits with
positive and negative $\pi/2$ shifts.
Each shifted circuit supplies 256 shots on MovingMNIST or 512 on HEP18.
The 32 unshifted predictions provide the independent reference samples
used to evaluate the shifted costs.
Across the full training schedule, the recorded budgets are approximately
144.6 million shots per MovingMNIST model and 566.5 million per HEP18 model.
These totals include both shifted measurements and unshifted training samples.
The experiments establish a measurement-based route to training the
prior; practical hardware training would also require reducing this sampling cost.

\begin{table}[tbp]
\centering\small
\setlength{\tabcolsep}{3pt}
\caption{MovingMNIST: quantum-training comparison. Mean $\pm$ sample standard deviation of three matched trainings. Coverage and validity are percentages; ADE and ES are in pixels. Noise denotes 0.5\% independent bit flips at training and evaluation. All settings use 128 evaluation samples per observation.}
\label{tab:shift-mnist}
\begin{tabular}{@{}lrrrrrr@{}}
\toprule
Training & \multicolumn{1}{c}{Cov. $\uparrow$} & \multicolumn{1}{c}{Valid $\uparrow$} & \multicolumn{1}{c}{TV $\downarrow$} & \multicolumn{1}{c}{minADE $\downarrow$} & \multicolumn{1}{c}{ES $\downarrow$} & \multicolumn{1}{c}{Dice $\uparrow$} \\
\midrule
\multicolumn{7}{@{}l}{\textbf{Product-QCBM}} \\
Ideal gradient & $90.9\,\pm\,5.1$ & $95.6\,\pm\,1.9$ & $0.179\,\pm\,0.023$ & $0.92\,\pm\,0.14$ & $9.64\,\pm\,0.03$ & $0.591\,\pm\,0.014$ \\
Parameter shift & $94.2\,\pm\,2.7$ & $96.4\,\pm\,1.6$ & $0.168\,\pm\,0.018$ & $0.80\,\pm\,0.07$ & $9.60\,\pm\,0.02$ & $0.601\,\pm\,0.005$ \\
Shift + noise & $94.3\,\pm\,2.4$ & $96.2\,\pm\,1.6$ & $0.167\,\pm\,0.019$ & $0.80\,\pm\,0.08$ & $9.60\,\pm\,0.02$ & $0.602\,\pm\,0.007$ \\
\midrule
\multicolumn{7}{@{}l}{\textbf{Entangled-QCBM}} \\
Ideal gradient & $93.1\,\pm\,2.3$ & $96.6\,\pm\,0.9$ & $0.159\,\pm\,0.016$ & $0.85\,\pm\,0.11$ & $9.60\,\pm\,0.03$ & $0.597\,\pm\,0.013$ \\
Parameter shift & $93.5\,\pm\,2.8$ & $95.9\,\pm\,1.6$ & $0.165\,\pm\,0.032$ & $0.84\,\pm\,0.11$ & $9.61\,\pm\,0.04$ & $0.598\,\pm\,0.012$ \\
Shift + noise & $94.0\,\pm\,2.3$ & $96.5\,\pm\,1.1$ & $0.160\,\pm\,0.024$ & $0.84\,\pm\,0.11$ & $9.61\,\pm\,0.03$ & $0.595\,\pm\,0.014$ \\
\bottomrule
\end{tabular}
\end{table}

\begin{table}[tbp]
\centering\small
\setlength{\tabcolsep}{3pt}
\caption{HEP18: quantum-training comparison. Mean $\pm$ sample standard deviation of three matched trainings. Coverage and validity are percentages; ADE and ES are in metres. Noise denotes 0.5\% independent bit flips at training and evaluation. All settings use 128 evaluation samples per observation.}
\label{tab:shift-intersection}
\begin{tabular}{@{}lrrrrr@{}}
\toprule
Training & \multicolumn{1}{c}{Cov. $\uparrow$} & \multicolumn{1}{c}{Valid $\uparrow$} & \multicolumn{1}{c}{TV $\downarrow$} & \multicolumn{1}{c}{minADE $\downarrow$} & \multicolumn{1}{c}{ES $\downarrow$} \\
\midrule
\multicolumn{6}{@{}l}{\textbf{Product-QCBM}} \\
Ideal gradient & $39.45\,\pm\,10.31$ & $25.56\,\pm\,4.21$ & $0.765\,\pm\,0.037$ & $4.155\,\pm\,0.530$ & $42.53\,\pm\,0.63$ \\
Parameter shift & $41.17\,\pm\,5.07$ & $25.80\,\pm\,4.46$ & $0.763\,\pm\,0.037$ & $4.332\,\pm\,0.482$ & $42.58\,\pm\,0.52$ \\
Shift + noise & $38.77\,\pm\,5.13$ & $24.50\,\pm\,4.11$ & $0.777\,\pm\,0.033$ & $4.230\,\pm\,0.490$ & $42.45\,\pm\,0.53$ \\
\midrule
\multicolumn{6}{@{}l}{\textbf{Entangled-QCBM}} \\
Ideal gradient & $99.83\,\pm\,0.17$ & $98.55\,\pm\,0.40$ & $0.164\,\pm\,0.013$ & $0.189\,\pm\,0.019$ & $34.31\,\pm\,0.01$ \\
Parameter shift & $99.80\,\pm\,0.16$ & $96.91\,\pm\,2.18$ & $0.169\,\pm\,0.023$ & $0.187\,\pm\,0.016$ & $34.32\,\pm\,0.04$ \\
Shift + noise & $99.73\,\pm\,0.25$ & $93.38\,\pm\,6.04$ & $0.188\,\pm\,0.036$ & $0.199\,\pm\,0.023$ & $34.36\,\pm\,0.03$ \\
\bottomrule
\end{tabular}
\end{table}

\paragraph{MovingMNIST results.}
Both circuit families retained high validity and similar frequency
accuracy when trained with parameter shift.
Some displayed means improved slightly over ideal-gradient training,
but three runs do not establish a benefit from adding noise.

\paragraph{HEP18 results.}
Noise-free parameter shift retained nearly complete coverage and similar
trajectory errors for Entangled-QCBM.
The noisy setting had mean joint validity of 93.38\%, compared with
98.55\% for the ideal-gradient reference.
Its between-training standard deviation increased to 6.04 percentage points.
The three noisy runs ranged from 86.41--97.09\% validity, while each
recovered more than 99\% of the joint future modes.

Product-QCBM had substantially lower joint validity under all three
training settings.
Thus, Entangled-QCBM continued to learn useful cross-agent correlations
when gradients were estimated from measurements.
The noisy model nevertheless had lower mean validity than the strongest
classical correlated priors in the main comparison.

\section{External Learning-Method Implementations}
\label{app:external}
These comparisons test different ways of learning a predictive distribution.
They complement the prior comparisons in \cref{sec:comparison-methods},
which keep our sampled predictive objective fixed.
All models receive observations and realized futures, without action or
mode labels.
We retain the same benchmark encoder architectures and 64-dimensional
representations.
The external prediction networks use hidden width $H=96$ for
MovingMNIST and $H=64$ for HEP18.
For HEP18, the encoder concatenates four 16-dimensional agent representations.
External predictors receive the complete observation representation;
we do not impose our agent-local code restriction on these baselines.

\paragraph{Variational JEPA visual adaptation.}
We use the training objective from the authors' public MovingMNIST
visual implementation of Variational JEPA \cite{huang2026varjepa}.
An online encoder produces the observation representation $h$.
A second encoder, updated by exponential moving average (EMA), produces
the target representation $t$.
Separate $d\!\rightarrow\!H\!\rightarrow\!d$ SiLU MLPs predict the
mean $\mu(h)$ and log variance $\ell(h)$ of a Gaussian future representation.
We clamp the predicted log variances to the range $[-10,10]$.

We train the mean prediction using VICReg between $\mu(h)$ and the
detached target.
The MSE term has weight 25, each branch's variance-hinge term has
weight 25, and the off-diagonal covariance term has weight 1.
The variance head uses the Gaussian calibration loss
\begin{equation*}
 \mathcal L_{\rm var}
 =\frac12\E\!\left[\sum_{j=1}^{d}
 \ell_j+\frac{(t_j-\mu_j)^2}{\exp(\ell_j)}\right].
\end{equation*}
We give this term weight 0.1 and detach the observation representation,
mean prediction and target when computing it.
It therefore updates only the variance head, which learns the
uncertainty around the mean prediction.
At inference, we draw future representations from
$\widehat t\sim\mathcal N(\mu(h),\operatorname{diag}(\exp\ell(h)))$.
This adaptation follows the visual implementation's VICReg and variance-calibration
training procedure, rather than reproducing every objective considered in the paper.

We use the benchmark-specific encoder and readout to accommodate each
task's input and output, replacing the original single-frame CNN.
Training uses Adam with learning rate $5\times10^{-4}$ and global
gradient clipping at norm 5.
We update the EMA encoder after every optimizer step, using coefficient
0.99 for MovingMNIST and 0.999 for HEP18.
Development evaluation selected these settings from the two coefficients
and the original once-per-pair-epoch update schedule.
A pair epoch counts observation--future pairs, including all target
views when a batch groups several futures under one observation.
We train for 4,000 updates on MovingMNIST and 6,000 on HEP18,
without test-selected early stopping.

\paragraph{Conditional variational latent-state baseline.}
Latent-CVAE uses one trainable encoder for both observations and targets.
Its posterior $q_{\omega}(z\mid h,t)$ sees the observation and realized
future during training.
Its prior $p_{\eta}(z\mid h)$ sees only the observation and is used
for sampling at inference.
Both distributions are diagonal Gaussians over $z\in\mathbb R^{12}$.
We parameterize them with
$128\!\rightarrow\!H\!\rightarrow\!24$ and
$64\!\rightarrow\!48\!\rightarrow\!24$ SiLU MLPs, respectively.
Each network's 24 outputs split into 12 means and 12 log variances,
with log variances clamped to $[-10,10]$.
During training, we sample the posterior through the reparameterization
$z=\mu_q+\exp(\ell_q/2)\odot\varepsilon$,
$\varepsilon\sim\mathcal N(0,I)$, and predicts
$\widehat t=h+g([h;z])$, where $g$ is a
$76\!\rightarrow\!H\!\rightarrow\!64$ SiLU MLP.
The objective is
\begin{equation*}
 \frac1d\E\|\widehat t-\sg(t)\|_2^2
 +\frac{\beta}{12}\E\,D_{\rm KL}(q_{\omega}\|p_{\eta})
 +\frac43\,\frac{R(h)+R(t)}2.
\end{equation*}
The first term trains the prediction to match the encoded future.
The KL term encourages the observation-only prior to match the
future-conditioned posterior.
VISReg regularizes both state representations using the implementation
in \cref{app:visreg}.
The reconstruction target is detached, but gradients can still reach
the target encoder through the posterior and regularization paths.
At inference, we sample only from $p_{\eta}(z\mid h)$, with no target input.

We use Adam at learning rate $10^{-3}$, global gradient clipping at
norm 5, and the same update budgets as Variational JEPA.
Development evaluation over $\beta\in\{0.01,0.1,1,3,10\}$ selected
$\beta=1$ for both tasks.

\paragraph{Readout and evaluation.}
For both baselines, we freeze the complete world model before fitting
the benchmark-specific readout on encoded real futures.
Variational JEPA supplies targets from its EMA encoder; Latent-CVAE uses
its shared encoder.
At inference, each readout decodes sampled predicted representations instead.
We train readouts for 4,000 updates on MovingMNIST and 16,000 on HEP18.
The data splits, decoded-output validity thresholds and sample counts
match the within-framework comparisons.
Each baseline is trained three times from independent initializations.
Reported standard deviations describe variation across these trained
models, not variation among a single model's 128 draws.
\begin{table}[tbp]
\centering\small
\setlength{\tabcolsep}{5pt}
\caption{Trainable world-model parameters in the two external-method
adaptations, reported separately for each dataset.
Counts include the encoder and the networks that predict future
representations or their distributions.
They exclude the non-trainable EMA encoder in Variational JEPA and
the separately fitted readout in both methods.}
\label{tab:external-parameters}
\begin{tabular}{@{}llr@{}}
\toprule
Task & Method & \multicolumn{1}{c}{Trainable parameters} \\
\midrule
MovingMNIST & Variational JEPA & 464,496 \\
MovingMNIST & Latent-CVAE & 472,208 \\
HEP18 & Variational JEPA & 22,288 \\
HEP18 & Latent-CVAE & 28,848 \\
\bottomrule
\end{tabular}
\end{table}

\subsection{Interpreting the complete-method comparison}
\label{app:external-analysis}
The improvements shared by classical and quantum priors in
\cref{tab:mnist,tab:intersection} support the evaluated sampled
latent-action training recipes. These comparisons change several
training choices and do not isolate an individual loss or regularizer.
Their conclusions apply to the adaptations and budgets specified above.

Best-sample accuracy and distributional validity need not agree.
Latent-CVAE has higher MovingMNIST coverage and lower minimum ADE,
yet many draws fall between the valid directions. On HEP18, its energy
score is close to ours despite much lower joint validity: incompatible
agent combinations can still have moderate average trajectory error.
Variational JEPA also has low joint validity, although its readout
accurately decodes encoded real targets, pointing to errors in predicted
features rather than target reconstruction.

\section{Experimental Protocol Details}
\label{app:protocol}
\subsection{Common training and comparison methods}
\label{app:comparison-methods}
Section~\ref{sec:comparison-methods} defines the comparison methods.
Within each stochastic discrete-prior comparison, models are trained
separately with the same encoder, predictor, energy loss and schedule.
Appendix~\ref{app:training} gives their network and optimizer settings;
Appendix~\ref{app:circuits} specifies the circuits and parameter counts.
The external learning methods replace the objective as detailed in
Appendix~\ref{app:external}, while retaining the benchmark data,
encoder architectures, readouts and evaluation budgets.

For MovingMNIST and HEP18, we evaluate three independent training runs
per method, using 128 prior draws per observation and retaining repeated
codes. Deterministic JEPA repeats its single continuation 128 times
under the same metrics. The trained QCBMs use noisy parameter shift;
Frozen-QCBM keeps its initialized, noise-free prior. D3IL's distinct
training and aggregation protocol is specified in
Appendix~\ref{app:d3il-implementation}.
Table shading identifies trained QCBMs; bold and underlined entries
mark the best and second-best displayed means, not statistical significance.

\paragraph{Simulation and readout noise.}
We run the neural networks and PennyLane circuit sampler on an RTX~5090.
This paragraph describes the reference experiments;
Appendix~\ref{app:d3il-implementation} specifies the demonstration study.
The simulator evolves a statevector internally, but the parameter-shift
update receives only measured bit strings. It estimates gradients from
samples, without differentiating the simulated quantum probabilities.
Circuit simulation uses double precision for MovingMNIST and single
precision for HEP18.

In the noisy setting, we independently flip each measured bit with
probability 0.005, during both training and evaluation.
The noise-free parameter-shift reference uses the same procedure without
these flips. This noise model covers readout flips only; it does not
include gate errors, decoherence or device-specific calibration noise.

\subsection{Stochastic MovingMNIST}
\label{app:mnist}
We construct single-digit videos from MNIST sprites using the MovingMNIST
setting \cite{srivastava2015unsupervised}. Each observation contains three
$64\times64$ frames showing the same digit moving horizontally.
The digit centre passes through $x=18,21,24$ pixels at a fixed vertical
coordinate drawn uniformly from integers 24--40.
Each future contains three more frames, with a displacement of four
pixels per frame.
The future direction is one of eight angles, separated by $45^\circ$
and starting from horizontal rightward motion in image coordinates.
The four cardinal directions each have probability 0.15; the four
diagonal directions each have probability 0.10.
A separate digit-dependent probability experiment appears in
Appendix~\ref{app:conditional-prior}; it does not replace this reference task.

We use 800 training observations from MNIST's training split and 80 test
observations from its test split.
We reserve another 80 observations from the MNIST training split for
development, disjoint from the 800 fitting observations. Development
first checks readout viability and then compares decoded trajectory
energy score; the official test split is not used for this selection.
Each training observation has 20 records in permuted order, comprising
three continuations per cardinal direction and two per diagonal direction.
These frequencies implement the specified 0.15 and 0.10 direction probabilities.
Each training example supplies the observation images and one realized
future clip. Direction labels, centre coordinates and the list of
possible futures are used only for evaluation, not as model inputs.
The three-bit register provides eight possible codes, whose effects on
motion are learned. Multiple codes may learn similar continuations;
we do not impose a one-to-one assignment.

The target probabilities can be represented by independent bits under
a suitable learned assignment of codes to directions.
One bit with probability 0.6 and two fair bits give four code
probabilities of 0.15 and four of 0.10.
We do not provide this assignment to the model during training.
The benchmark therefore tests whether the model learns distinct futures
and their frequencies; it does not require entanglement.

The shared encoder maps each three-frame clip to a representation.
The predictor combines the observation representation with a three-bit
action to predict a future representation.
Appendix~\ref{app:networks} specifies the encoder, predictor and readout architectures.

\paragraph{Frozen translation-based image readout.}
We train the image readout after freezing the encoder, predictor and prior.
During readout training, its input is the concatenation of the observation
and target representations, $[h;t]$.
At inference, we replace the encoded target with the predicted
representation and supply $[h;\widehat t]$ instead.
The readout predicts one two-dimensional displacement for each future frame.
A bilinear spatial transformer shifts the last observed image by these
displacements to produce the future clip.

This design uses the benchmark's translation structure, so the learned
readout estimates motion rather than synthesizing digit appearance from scratch.
Its image loss updates only the readout, leaving all JEPA parameters fixed.
Every method uses the same readout architecture, training budget and
target-feature training procedure, detailed in Appendix~\ref{app:networks}.

\subsection{HighwayEnv priority intersection (HEP18)}
\label{app:intersection}
We construct HEP18 on the straight intersection lanes of HighwayEnv
\cite{leurent2018highwayenv}, with four approaching vehicles and 18 possible joint futures.
The vehicles approach the intersection from north, west, south and east.
For each vehicle, we sample its initial speed uniformly over
$[8,9]$\,m/s and its distance from the centre over $[43,47]$\,m.
The observation records two seconds of motion at these constant speeds.
The future spans nine seconds, divided into three stages of three seconds each.
Within each stage, acceleration is constant and we integrate the motion exactly.
We use HighwayEnv's road geometry with these specified motion schedules,
rather than its default driver policy.

For each observation, either the north--south (NS) or east--west (EW)
axis has priority, with probability one half.
Each vehicle on the priority axis independently chooses one of three
equally likely acceleration schedules:
\begin{equation}
 A=(-0.5,0,0.5),\quad B=(0,0,0),\quad C=(0.5,0,0)
 \quad\text{in m/s$^2$.}
 \label{eq:schedules}
\end{equation}
The entries specify acceleration during the first, second and third stages.
Schedule $A$ decelerates, maintains speed and then accelerates; $B$ maintains
speed throughout; $C$ accelerates first and then maintains speed.
Both yielding vehicles use the braking schedule
$D=(-v_0/5,-v_0/10,-v_0/30)$ in m/s$^2$, where $v_0$ is their
initial speed in m/s and each coefficient has units s$^{-1}$.
Their speeds after the three stages are $0.4v_0$, $0.1v_0$ and zero,
respectively. The two priority choices and three schedules per priority
vehicle give $2\times3^2=18$ equally probable joint futures.
The initial-state ranges let yielding vehicles stop before the intersection
and let priority vehicles clear it even under schedule $A$.
During construction, we check vehicle rectangles for overlap at intervals
of 0.05\,s along every trajectory.

The dataset contains 448 observations, with 256 for training, 64 for
validation and 128 for testing.
We split by observation, keeping all 18 associated futures together.
We choose the final HEP18 training configuration using only the
64 validation observations.
Each final checkpoint is then evaluated on the separate 128-observation
test split, without further parameter updates.

Each stored state contains
$(x/50,y/50,v_x/10,v_y/10,\cos\phi,\sin\phi)$.
Here $\phi$ is the vehicle heading, and positions and velocities are
scaled as shown. For learning, we project position and velocity onto
each vehicle's known straight route.
Each agent is then described by three longitudinal position--speed pairs.
Observation times are $-2,-1,0$\,s, and future times are $3,6,9$\,s.
Both views therefore contain six scalar inputs per agent, despite covering
different time intervals.

The same per-agent encoder maps both views to local representations.
We concatenate the four agent representations to describe the complete scene.
The predictor learns a residual change in these learned representation coordinates.
Training uses the structured states, while schedule labels are reserved
for evaluation. Appendix~\ref{app:networks} gives the network dimensions.

\paragraph{Agent-local latent actions.}
We divide the twelve action bits into four groups of three, one per vehicle.
This lets us test whether correlations in the prior coordinate the
vehicles' predicted futures.
Each agent's predictor receives the full observation representation and
its own three random bits. Before the shared code MLP, a mask sets
the other agents' signed-bit inputs to zero.
The predictor updates only that agent's 16-dimensional representation,
and the trajectory readout also decodes each agent separately.
For a fixed observation, cross-agent stochastic coordination must therefore
come from correlations among the prior's bit groups.

\paragraph{Joint predictive energy.}
We evaluate one energy score on the concatenated four-agent representation.
A prediction is therefore compared with a complete target scene, rather
than matching each vehicle to a different target future.
This makes the loss sensitive to coordination between vehicles as well
as their individual motion. For representation dimension $d$, we use
\begin{equation}
 \rho_{\beta,\epsilon}(u,v)
 =\left(\frac{\|u-v\|_2^2}{d}+\epsilon^2\right)^{\beta/2}
       -\epsilon^\beta,\qquad \beta=0.5,\quad\epsilon=10^{-4}.
 \label{eq:power-distance}
\end{equation}
With $\beta=0.5$, the distance grows sublinearly, giving more relative
weight to differences between nearby representations.
The positive $\epsilon$ keeps the derivative smooth when two representations coincide.
We average the prediction--target term over all 18 encoded futures
associated with the observation.
Both the neural-network update and the prior update use this distance.
For evaluation in physical coordinates, we use the ordinary Euclidean
energy score instead.

\paragraph{Change regularization.}
For HEP18, we additionally regularize the change representation with
$\lambda_{\Delta}R(t-h)$, where $\lambda_{\Delta}=2/3$.
Here $t-h$ measures the change from the observed representation to a
realized future representation.
Regularizing it encourages the encoder to preserve variation in how
the scene continues. The constraint acts on learned features; it does
not assume that physical changes in position or speed follow a Gaussian distribution.
We compute $R(t-h)$ for each of the 18 target views and average the results.
Together with \cref{eq:regviews}, this gives a coefficient of $2/3$ on
each of $R(h)$, $R(t)$ and $R(t-h)$.
MovingMNIST uses only the observation and target regularizers, $R(h)$ and $R(t)$.
Appendix~\ref{app:objective-stability} tests the effect of change regularization
while keeping the predictive objective and per-state weights fixed.

\paragraph{Frozen trajectory readout.}
We train the trajectory readout after freezing all JEPA parameters.
A shared per-agent MLP maps an encoded real future to three predicted
speeds, one at each future time.
At inference, it receives the model's predicted representation in place
of the encoded real future.
Starting from the observed position and speed, we integrate these speeds
with the trapezoidal rule at three-second intervals.
The MLP receives no observation features; the observed state only initializes
this fixed integration step.

We fit a separate readout for each trained model using encoded real futures.
Readout training never changes the encoder, predictor or prior.
Appendix~\ref{app:networks} specifies the readout architecture, loss and optimization settings.

\subsection{Prediction quality and multi-agent coordination metrics}
\label{app:metric-definitions}
The reference benchmark metrics use 128 prior samples per observation, including
repeated codes. We decode these samples before evaluating prediction quality.

\paragraph{MovingMNIST validity and coverage.}
We assign each prediction to the true direction with the nearest final endpoint.
The sample is valid if this endpoint error is less than 4 pixels;
otherwise, we assign it to an invalid category.
Coverage is the fraction of the eight directions reached by at least
one valid sample. Validity is the fraction of all 128 samples that
pass the endpoint test.
These metrics answer different questions: coverage measures how many
alternatives are recovered, whereas validity measures how reliably the prior samples plausible futures.

\paragraph{HEP18 joint validity and coverage.}
A joint prediction must match one complete four-agent target future.
At each of the three future times, every agent must have position
error at most 2\,m and speed error at most 1\,m/s.
For each candidate target, we divide the errors by their respective
tolerances and take the largest value across agents and times.
We assign the prediction to the target with the smallest resulting error.
If this error exceeds one, the sample enters the invalid category.
Coverage is the fraction of the 18 targets matched by at least one
sample; validity is the fraction of valid joint samples.
The target identities are used for evaluation only, without supplying
mode labels during training.

\paragraph{Frequency accuracy.}
Total variation (TV) compares the sampled category frequencies with the
specified target probabilities, including the invalid category.
It equals one half of the sum of absolute frequency differences
across categories. MovingMNIST has eight direction categories plus one
invalid category; HEP18 has 18 joint-future categories plus one invalid category.
The target probability of the invalid category is zero for both tasks.
MovingMNIST uses the specified cardinal and diagonal probabilities, while
HEP18 assigns probability $1/18$ to each valid future.
Lower TV indicates a closer match to these probabilities, accounting
for both invalid samples and incorrect frequencies among valid futures.

\paragraph{Trajectory and image accuracy.}
For minimum average displacement error (minADE), we find the sampled
rollout with the smallest mean position error for each true future.
Minimum final displacement error (minFDE) instead uses the position error
at the last future time.
Each metric selects a complete rollout, rather than combining different
samples across agents or time steps.
We average over target futures using their specified probabilities:
the cardinal/diagonal weights for MovingMNIST and uniform weights for HEP18.

The trajectory energy score (ES) uses Euclidean distances between flattened
complete trajectories, measured in pixels for MovingMNIST and metres for HEP18.
It evaluates the sampled distribution, rather than only the closest
prediction to each target.
For MovingMNIST, we also report the best-sample image Dice score,
averaged over true futures with their specified probabilities.
The Dice score measures overlap between predicted and target digit images.
When evaluating the readout alone, we replace predicted features with
encoded real targets. This diagnostic separates readout reconstruction
quality from errors introduced by the world model.

\paragraph{Cross-agent coordination.}
For each observation, we independently shuffle the sampled trajectories
of each vehicle and recombine them into four-agent scenes.
This preserves every vehicle's set of trajectories and frequencies, while
breaking their original association with other vehicles.
We recompute joint validity and average the result over ten shuffles.
A drop in validity shows that the original pairing contained useful
cross-agent coordination.
Separately, local validity checks whether each vehicle matches any of
its four allowed schedules, without requiring a compatible four-agent combination.
\Cref{app:metrics} reports checks with different sampling budgets and validity tolerances.

\begin{figure}[!t]
\centering
\includegraphics[width=\textwidth]{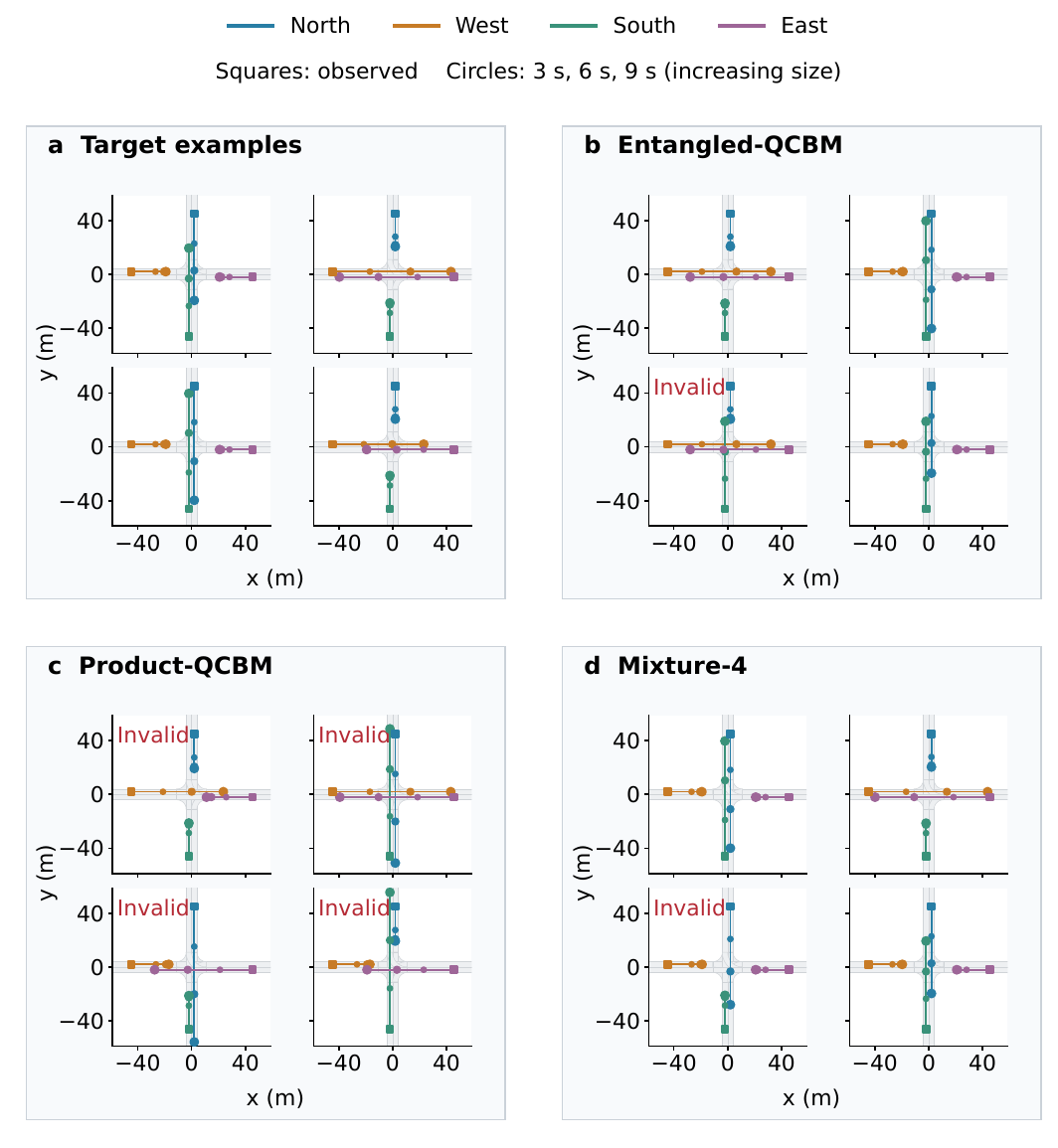}
\caption{Diverse continuation examples in HEP18, using ideal-gradient reference checkpoints.
Each group shows four examples for the same test observation, with
targets in (a) and sampled predictions in (b--d).
We select distinct combinations of local behaviours from the 18 targets
or each model's 128 prior draws, favouring visibly different trajectories.
All groups use the first test observation and first training run.
Selection is independent across groups, so examples in corresponding
positions are not paired.
The display illustrates different continuations; their counts do not
estimate how often the model samples them.
Invalid predictions are labelled, and all groups share agent colours
and spatial limits. Increasing marker sizes indicate 3, 6 and 9\,s;
squares mark the observed positions.
The numerical tables use all prior draws, without this selection for display.}
\label{fig:intersection}
\end{figure}

\section{Change-Regularization Stability}
\label{app:objective-stability}
We test whether regularizing $t-h$ improves training stability on HEP18.
The alternative removes only $R(t-h)$, retaining the joint predictive
energy and the same coefficients on $R(h)$ and $R(t)$.
We train each configuration from three matched initializations and
evaluate it on the 64-observation validation split.
Quantum models use ideal circuit gradients in this ablation, so
finite-shot noise does not confound the objective comparison
(\cref{tab:training-stability}).
\begin{table}[H]
\centering\small
\setlength{\tabcolsep}{5pt}
\caption{Effect of change regularization on 64 HEP18 validation observations.
Values are means $\pm$ sample standard deviations across three matched
initializations, with 128 prior draws per observation.
All rows use joint predictive energy and coefficient $2/3$ on
each of $R(h)$ and $R(t)$.
The retained configuration also includes $\frac{2}{3}R(t-h)$.
Coverage and validity are percentages, and every run is retained,
including unsuccessful training outcomes.}
\label{tab:training-stability}
\begin{tabular}{@{}lcrrr@{}}
\toprule
Prior & \multicolumn{1}{c}{$R(t-h)$} & \multicolumn{1}{c}{Cov. $\uparrow$} & \multicolumn{1}{c}{Valid $\uparrow$} & \multicolumn{1}{c}{TV $\downarrow$} \\
\midrule
Entangled-QCBM & Yes & $99.88\,\pm\,0.05$ & $98.62\,\pm\,0.37$ & $0.166\,\pm\,0.012$ \\
Entangled-QCBM & No & $70.72\,\pm\,50.34$ & $67.47\,\pm\,52.74$ & $0.417\,\pm\,0.451$ \\
Mixture-4 & Yes & $99.80\,\pm\,0.05$ & $97.86\,\pm\,1.69$ & $0.157\,\pm\,0.004$ \\
Mixture-4 & No & $90.65\,\pm\,16.11$ & $91.91\,\pm\,10.37$ & $0.228\,\pm\,0.121$ \\
\bottomrule
\end{tabular}
\end{table}

Removing change regularization produced much greater variation across
training runs for Entangled-QCBM.
Two runs retained high coverage and validity, but the third generated
mostly invalid continuations.
For that run, decoding encoded real targets still gave 100\% validity.
The readout could therefore reconstruct valid motion from target features;
the failure arose when using the model's predicted features.
Mixture-4 also lost accuracy in its third run, although less severely.

With change regularization, all three Entangled-QCBM runs retained high
coverage and validity.
This ablation supports retaining $R(t-h)$ to stabilize the present
HEP18 training setup.
It evaluates ideal-gradient training; the separate finite-shot comparison
is reported in Appendix~\ref{app:gradient-comparison}.

\section{Evaluation Checks and Direction Frequencies}
\label{app:metrics}
The sampling-budget checks and direction-frequency figure
use ideal-gradient reference checkpoints, as indicated in their captions.

\paragraph{Interpreting the metrics.}
For HEP18, each sampled four-agent trajectory is assigned to its
nearest target using the normalized-error rule in Appendix~\ref{app:metric-definitions}.
If the sample exceeds the tolerance, it enters the invalid category.
A model can therefore have high coverage but low validity: it may
occasionally reach every target while producing many invalid samples.
Conversely, a model that repeatedly predicts one valid future can have
high validity but low coverage.
TV also measures whether the frequencies of valid futures are correct.
The metrics capture different properties and need not rank methods identically.
A deterministic model repeats the same continuation within each observation,
so shuffling its draws leaves all metrics unchanged.

\paragraph{Sampling-budget and tolerance checks.}
We reevaluate the first 32 and first 64 draws from each existing
128-draw sample set.
Using these nested subsets changes the sampling budget without drawing
a new set of predictions.
We also test position/speed tolerances of 1\,m/0.5\,m/s and
3\,m/1.5\,m/s, keeping all 128 draws.
The main evaluation retains its fixed tolerances of 2\,m and 1\,m/s.
These checks show how the reported metrics respond to stricter matching
or fewer samples.

For a finite-sample reference, we draw 128 futures directly from the
true uniform distribution over the 18 targets.
Its empirical TV is generally nonzero because sampled frequencies
fluctuate, even when the underlying distribution is correct.
\begin{table}[tbp]
\centering\small
\setlength{\tabcolsep}{5pt}
\caption{Ideal-gradient reference: sampling-budget and tolerance checks
on 128 HEP18 test observations.
Values are means over three independent training runs.
The sample-budget rows use the first $K$ draws from the same
128-draw set, with position/speed tolerances of 2\,m/1\,m/s.
The tolerance rows retain all 128 draws and change only the
matching thresholds.}
\label{tab:sensitivity}
\begin{tabular}{@{}llrrr@{}}
\toprule
Prior & Setting & \multicolumn{1}{c}{Coverage (\%)} & \multicolumn{1}{c}{Validity (\%)} & \multicolumn{1}{c}{TV} \\
\midrule
Entangled-QCBM & $K=32$ & 82.60 & 98.59 & 0.31 \\
Entangled-QCBM & $K=64$ & 96.88 & 98.53 & 0.22 \\
Entangled-QCBM & $K=128$ & 99.83 & 98.55 & 0.16 \\
Entangled-QCBM & 1 m / 0.5 m/s & 91.42 & 82.84 & 0.28 \\
Entangled-QCBM & 3 m / 1.5 m/s & 99.83 & 98.70 & 0.16 \\
Product-QCBM & $K=32$ & 25.77 & 25.44 & 0.80 \\
Product-QCBM & $K=64$ & 34.33 & 25.48 & 0.78 \\
Product-QCBM & $K=128$ & 39.45 & 25.56 & 0.77 \\
Product-QCBM & 1 m / 0.5 m/s & 24.68 & 10.78 & 0.89 \\
Product-QCBM & 3 m / 1.5 m/s & 44.24 & 27.96 & 0.74 \\
Mixture-4 & $K=32$ & 82.83 & 97.97 & 0.30 \\
Mixture-4 & $K=64$ & 96.70 & 97.94 & 0.22 \\
Mixture-4 & $K=128$ & 99.96 & 97.88 & 0.16 \\
Mixture-4 & 1 m / 0.5 m/s & 92.00 & 85.36 & 0.26 \\
Mixture-4 & 3 m / 1.5 m/s & 99.96 & 98.68 & 0.16 \\
AR-GRU & $K=32$ & 78.43 & 95.43 & 0.34 \\
AR-GRU & $K=64$ & 91.48 & 95.39 & 0.26 \\
AR-GRU & $K=128$ & 94.16 & 95.54 & 0.21 \\
AR-GRU & 1 m / 0.5 m/s & 81.05 & 75.07 & 0.35 \\
AR-GRU & 3 m / 1.5 m/s & 94.33 & 96.57 & 0.20 \\
\bottomrule
\end{tabular}
\end{table}

\paragraph{Run variation and repeated numerical entries.}
The $\pm$ values in the training-result tables are sample standard
deviations across independently trained models.
They do not describe variation among the 128 draws from one model.
Equal rounded entries can occur when different sample sets contain
the same codes needed to achieve the best predictions.
Further repetitions change code frequencies, but not coverage or
best-sample errors once those codes have been drawn.
Equal displayed values therefore do not imply identical sampled distributions.

\begin{figure}[!htbp]
\centering
\includegraphics[width=\textwidth]{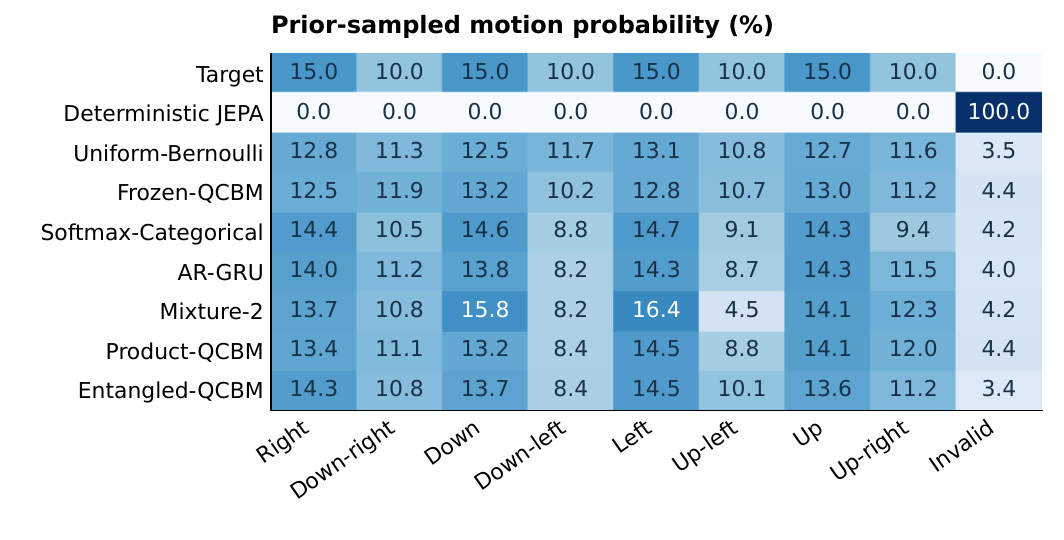}
\caption{MovingMNIST direction frequencies from the ideal-gradient reference models.
We average sampled frequencies over all test observations and three
independently trained models, retaining the invalid category.
Printed values give the actual percentages in every cell.
Colour intensity is capped at 25\%, so concentrated rows do not obscure
differences among the more evenly distributed rows.}
\label{fig:mnist-probabilities}
\end{figure}

\section{Learning Digit-Dependent Future Probabilities}
\label{app:conditional-prior}
We first train on data in which different digits have different
future-direction probabilities. We then test whether the trained prior
uses the observation by changing only its input features during evaluation.

\paragraph{Step 1: digit-dependent data and ordinary training.}
We use handwritten zeros and ones. Both retain all eight directions,
starting with rightward motion and advancing by $45^\circ$ in image
coordinates. Their direction probabilities are
\begin{align}
 q_0 &= (0.24,0.06,0.16,0.04,0.24,0.06,0.16,0.04),\\
 q_1 &= (0.125,0.125,0.125,0.125,0.125,0.125,0.125,0.125).
\end{align}
The direction distribution is non-uniform for digit 0 and uniform for digit 1.
Per digit, we use 400 training and 40 disjoint development sprites from
MNIST's training split, and 40 sprites from its official test split.
Vertical positions are paired across digit classes. The clips and motion
steps follow Appendix~\ref{app:mnist}. Each training example supplies
observation images and one sampled future. Digit labels only set the
sampling probabilities; the model is trained from images.

Both QCBMs use the three-bit MovingMNIST architectures, energy score,
two-view VISReg, optimizers and warm-up schedule in
Appendix~\ref{app:training}.
We train three models per circuit family, each with 4,000 JEPA updates
followed by 4,000 readout updates with the JEPA weights frozen.
We use the parameter-shift budgets
in \cref{tab:configuration}, with 0.5\% independent readout flips during
training and evaluation. The recorded budget is 145,536,000 simulated
training shots per run. We retain all six final checkpoints without
using the reserved development bank for tuning or checkpoint selection.
The ``Original observation'' rows in \cref{tab:conditional-mnist} report
normal prediction on this new dataset, averaged across both digits.

\paragraph{Step 2: testing the prior's use of the observation.}
Both the prior and predictor receive the observation, so either can
make the predicted direction distribution depend on the digit.
To isolate the prior's contribution, we fix all weights and keep the
original observation representation $h$ at the predictor and readout.
We replace only the prior input with $\widetilde h$:
$Z\sim p_\theta(\cdot\mid\widetilde h)$ and
$\widehat t=P_\psi(h,Z)$.
Normal inference uses $\widetilde h=h$.

We compare the original input with three replacements.
These are the next image of the same digit, the other digit at the same
vertical position, and a randomly selected training observation.
For the last condition, we draw a new training observation for every prior sample.
For example, a zero still enters the predictor while the prior receives
features of a one; predictions are still scored against $q_0$.
All conditions are scored against the original digit's target probabilities.
The random training inputs remove the link between the prior input
and the current observation. The same-digit replacement can also
change vertical position.
For each of 80 test observations, we pool three groups of 128 predictions,
compute TV including invalid samples, and average over observations.
We report standard deviations across the three independent training runs.

\begin{table}[tbp]
\centering\small
\setlength{\tabcolsep}{3pt}
\medmuskip=0mu
\caption{Digit-dependent MovingMNIST: mean $\pm$ sample s.d. over three training runs. Only the prior input changes between conditions; the predictor and readout keep the original observation. Each test observation has 384 sampled futures. TV includes invalid predictions; coverage and validity are percentages. Training uses finite-shot parameter shift and 0.5\% readout flips for both priors.}
\label{tab:conditional-mnist}
\begin{tabular}{@{}lrrrr@{}}
\toprule
Prior & \multicolumn{1}{c}{Prior input} & \multicolumn{1}{c}{TV $\downarrow$} & \multicolumn{1}{c}{Coverage $\uparrow$} & \multicolumn{1}{c}{Valid $\uparrow$} \\
\midrule
\rowcolor{quantumrow}
Product-QCBM & Original observation & $0.199\,\pm\,0.009$ & $83.02\,\pm\,1.42$ & $94.67\,\pm\,0.48$ \\
 & Other image, same digit & $0.210\,\pm\,0.007$ & $83.02\,\pm\,1.42$ & $94.72\,\pm\,0.58$ \\
 & Other digit & $0.219\,\pm\,0.019$ & $83.02\,\pm\,1.42$ & $94.63\,\pm\,0.58$ \\
 & Training-input mixture & $0.207\,\pm\,0.012$ & $83.02\,\pm\,1.42$ & $94.55\,\pm\,0.68$ \\
\midrule
\rowcolor{quantumrow}
Entangled-QCBM & Original observation & $0.170\,\pm\,0.014$ & $87.34\,\pm\,0.78$ & $95.02\,\pm\,1.93$ \\
 & Other image, same digit & $0.197\,\pm\,0.018$ & $87.34\,\pm\,0.78$ & $94.29\,\pm\,2.29$ \\
 & Other digit & $0.236\,\pm\,0.026$ & $87.34\,\pm\,0.78$ & $94.75\,\pm\,1.73$ \\
 & Training-input mixture & $0.199\,\pm\,0.021$ & $87.34\,\pm\,0.78$ & $94.36\,\pm\,2.00$ \\
\bottomrule
\end{tabular}
\end{table}

\paragraph{Results and interpretation.}
Replacing the prior input with the other digit increases prediction TV
in all three runs of both circuit families. The paired mean increase is
$0.020\pm0.014$ for Product-QCBM and $0.065\pm0.023$ for Entangled-QCBM.
For Entangled-QCBM, the training-input mixture worsens
TV in all three runs; the Product-QCBM effect is smaller and reverses slightly
in one run. Thus the prior contributes useful observation-dependent
probabilities, with a stronger effect for Entangled-QCBM in this check.

After averaging sampled code frequencies within each digit class,
the TV between the two classes is
$0.041\pm0.018$ for Product-QCBM and $0.074\pm0.028$ for Entangled-QCBM.
Every input condition samples all eight codes, and each code's predicted
future stays fixed because the predictor input is unchanged.
Coverage therefore stays the same, while the direction frequencies change.
These differences are computed within each trained model, because bit
codes need not have the same meaning across initializations. Decoding
encoded real targets gives mean endpoint errors of 0.66 pixels for both
families, below the four-pixel matching threshold.

These results show that the prior uses observation features to adjust
future probabilities. Replacements within the same digit class also
affect TV, indicating sensitivity to more than digit identity.

\section{Real-QPU Inference Details}
\label{app:qpu}
\paragraph{Hardware access and division of work.}
We used the LQCloud cloud service to access the MQ02 and AGate-100
quantum processors for MovingMNIST and HEP18, respectively.
The QPU generated latent-action bit strings, while the encoder,
conditioner, predictor and readout ran classically.
Each measurement shot denotes one circuit execution and returns one bit string.

\subsection{MovingMNIST deployment}
We deployed trained three-qubit Product-QCBM and Entangled-QCBM priors
on the MQ02 quantum processor.
Both use unchanged ideal-gradient reference weights,
rather than the noisy parameter-shift checkpoints in the main tables.
For each of the first eight MovingMNIST test observations, the encoder
and conditioner produce the circuit's nine rotation angles.
The Entangled-QCBM circuit contains two CNOT gates, while Product-QCBM omits them.
Both circuit families include their trained terminal rotations.

Before submission, we checked that each exported circuit reproduced
the simulator's ideal probabilities.
We then used three basis-state preparations to verify the ordering
of bits returned by the device.
For each circuit, we collected 1,024 measurement shots without readout correction.
Each measured code was passed directly through the frozen predictor
and image readout.

The primary comparison uses the first 128 shots per observation,
matching the main evaluation's sample budget.
We divide the remaining shots into seven further groups of 128.
These groups measure repeated sampling of the same trained circuit,
not variation across independent training runs.

\Cref{tab:qpu} shows that hardware-generated actions can drive the complete
prediction pipeline with similar results to ideal sampling.
Within each circuit family, coverage and minimum ADE
were identical for the two sampling sources.
Both sources recovered all eight codes for these observations, so
they supplied the same set of possible decoded predictions.
Their code frequencies differed, which changed TV and energy score.
This three-qubit experiment evaluates hardware deployment; the larger-register
resource analysis is given in \cref{sec:resource-crossover}.
\begin{table}[tbp]
\centering\small
\setlength{\tabcolsep}{5pt}
\caption{Real-QPU inference on the first eight MovingMNIST test observations, using the first trained model of each circuit family. GPU and QPU rows both use 128 actual draws per observation; each circuit-family pair shares its frozen neural networks. Coverage and validity are percentages; ADE and ES are in pixels. These are single-model deployment results, not means across independent trainings.}
\label{tab:qpu}
\begin{tabular}{@{}llcccccc@{}}
\toprule
Prior & Sampling & Cov. $\uparrow$ & Valid $\uparrow$ & TV $\downarrow$ & minADE $\downarrow$ & ES $\downarrow$ & Dice $\uparrow$ \\
\midrule
Product-QCBM & GPU / ideal & 89.06 & 98.44 & 0.1678 & 1.059 & 9.633 & 0.567 \\
Product-QCBM & QPU & 89.06 & 98.73 & 0.1674 & 1.059 & 9.670 & 0.567 \\
\midrule
Entangled-QCBM & GPU / ideal & 96.88 & 96.48 & 0.1482 & 0.680 & 9.565 & 0.607 \\
Entangled-QCBM & QPU & 96.88 & 97.17 & 0.1520 & 0.680 & 9.548 & 0.607 \\
\bottomrule
\end{tabular}
\end{table}

\subsection{HEP18 deployment with equivalent circuit layouts}
\begin{table}[tbp]
\centering\small
\setlength{\tabcolsep}{5pt}
\caption{HEP18 deployment on AGate-100: one frozen 12-qubit Entangled-QCBM checkpoint and eight observations held out from layout selection. Values average all eight disjoint 128-draw groups per observation (1,024 draws total), including the matched ideal reference. All QPU layouts use the same angles and 29 CZ gates. Coverage and validity are percentages; ADE and ES are in metres.}
\label{tab:hep18-qpu-optimized}
\begin{tabular}{@{}lccccc@{}}
\toprule
Sampling / layout & Cov. $\uparrow$ & Valid $\uparrow$ & TV $\downarrow$ & minADE $\downarrow$ & ES $\downarrow$ \\
\midrule
Ideal sampling & 99.91 & 98.41 & 0.166 & 0.218 & 34.38 \\
\midrule
QPU: original layout & 99.05 & 76.88 & 0.308 & 0.238 & 34.99 \\
QPU: selected layout & 98.78 & 81.53 & 0.280 & 0.241 & 34.86 \\
\rowcolor{quantumrow}
QPU: reflected layout & 99.57 & 86.69 & 0.239 & 0.227 & 34.66 \\
\bottomrule
\end{tabular}
\end{table}

\paragraph{Why the hardware layout matters.}
A hardware layout assigns the circuit's qubits to physical qubits and
connections on the processor.
Gate and measurement errors vary across the device, so layouts implementing
the same ideal circuit can produce different sampled code frequencies.

\paragraph{Frozen model and circuit compilation.}
We used one fixed 12-qubit Entangled-QCBM checkpoint trained with the
noisy parameter-shift configuration in \cref{tab:intersection}.
The encoder, conditioner, predictor, readout and every circuit parameter
remained unchanged throughout deployment.
The four-layer circuit contains 60 single-qubit rotations, including
the terminal rotations, and 20 logical CNOT gates.

On the AGate-100 quantum processor, we mapped the twelve circuit qubits
to a $4\times3$ physical grid.
Three SWAP gates moved quantum states between connected qubits to
implement the required interactions.
The compiled circuit used 29 two-qubit controlled-Z (CZ) gates
and 67 single-qubit instructions.
After measurement, we restored the logical bit order before passing
codes to the predictor.
Numerical checks found absolute differences below $10^{-16}$ between
the ideal output probabilities before and after compilation.

\paragraph{Layout selection.}
We screened six physical layouts using device calibration information
and simulated sensitivity to individual injected gate errors.
We selected the layout with the lowest joint TV on four development
observations, then confirmed its improvement on 16 additional observations.
Next, we tested four reflected orientations of the grid using the
same four development observations.
We fixed the best orientation before evaluating the final observations.

These layout changes leave circuit angles, gate counts and the ideal
Born distribution unchanged.
The final assignment of virtual wires (the numbered qubits in the
exported circuit) to physical qubits was
$(67,66,3,68,65,36,69,64,37,70,63,38)$.
The logical-to-wire permutation was
$(0,1,2,6,7,8,9,10,11,3,4,5)$.
Its entries specify the virtual wire used by each logical qubit;
both lists use zero-based indices.

\paragraph{Final measurement protocol.}
The final comparison used eight further HEP18 test observations.
None was used for layout selection or the intermediate confirmation.
For each observation, we ran the original, selected and reflected
layouts with 1,024 shots each.
We rotated submission order across observations, completing and saving
each job before submitting the next.

Twelve binary-index basis preparations verified bit order across the
three layouts.
These controls prepare known patterns of zeros and ones, so returned
bits can be matched to their intended logical positions.
Six final all-zero/all-one controls completed the 42-job protocol.
We used no readout correction, dynamical decoupling, sample rejection
or model updates.
Every measured code entered the same frozen prediction pipeline.

\paragraph{Aggregation and results.}
For each observation and layout, we divide the 1,024 shots into eight
disjoint groups of 128.
We compute metrics within each group, then average over all groups
and observations in \cref{tab:hep18-qpu-optimized}.
The ideal reference uses the same observations, grouping and sample budget.
Thus each metric evaluation retains the 128-draw budget used in the
main experiments.
All evaluations use the position and speed tolerances in
Appendix~\ref{app:metric-definitions}.
The repeated groups characterize sampling from one checkpoint, not
variation across independently trained models.

The reflected layout increased validity by 9.81 percentage points
relative to the original layout.
It also reduced joint TV for each of the eight observations.
Its joint validity reached 86.69\%, compared with 98.41\% under ideal sampling.
All individual-agent continuations remained locally valid, but 13.31\%
of joint samples combined local actions that were incompatible with the same priority choice.
Independently re-pairing agents reduced validity to 12.95\%.
The original measured samples therefore retained substantial cross-agent
coordination despite the remaining hardware errors.

Using only the first group of 128 shots gave validity of 75.00\%,
81.35\% and 83.59\% for the original, selected and reflected layouts.
The headline averages retain every group, rather than selecting the
most favourable one.
This is a matched comparison of layouts on one device; it does
not measure robustness across different days.

\paragraph{Other deployment checks.}
We also tested readout bit-flip averaging and equivalent changes to
the circuit's phase frame.
Further tests split rotations into half-angle gates or executed CZ
gates sequentially instead of concurrently.
These alternatives did not consistently improve development results
and were omitted from the final setup.
An earlier confirmation batch was interrupted because some cloud-task
records became unavailable.
We retained that batch separately and did not combine it with
the complete final comparison.

\subsection{D3IL common-start recursive inference}
\label{app:d3il-qpu}
\paragraph{Deployment.}
We deployed the eight-qubit Product-QCBM and Entangled-QCBM checkpoints
from the first fold and training run of \cref{tab:d3il}.
Both were trained with noisy parameter shift and kept fixed during deployment.
Each prior generated 40 raw rollouts from the common stationary point
using the re-encoding and stopping rules in Appendix~\ref{app:d3il-metrics}.
Every latent action came from a fresh QPU measurement, including after
re-encoding a predicted segment. The neural networks ran on the GPU.

We used an eight-qubit chain on AGate-100, selected from device
calibration without target trajectories or predictive scores.
CNOT-to-CZ compilation and single-qubit gate fusion gave eight local
gate blocks for Product-QCBM, and 36 local blocks plus 14 CZs for
Entangled-QCBM. Noiseless compilation checks preserved Born probabilities
to within $4.22\times10^{-15}$ for both circuits.
Four known basis patterns, each measured with 128 shots, verified bit order.
Identical circuits within a recursive step shared a submission, with
a separate measurement shot for each path. We used raw measurements without readout
correction, dynamical decoupling or additional simulated noise.
Product-QCBM and Entangled-QCBM supplied 458 and 437 latent actions,
respectively. Together with the controls, this required 775 tasks and
1,407 measurement shots.
We checked returned instructions, mappings and shot counts, and replayed
the saved measurements to reproduce all 80 paths exactly.

\paragraph{Evaluation budget and metrics.}
We use the same geometric checks, route definitions and uniform
24-route reference as in \cref{tab:d3il}.
That table summarizes nine models per method, each evaluated with
384 raw rollouts, and reports expected metrics for a subset of
60 accepted paths.
The hardware check uses one checkpoint per prior and 40 raw rollouts,
because each recursive step requires a new conditional circuit execution.
Both accepted pools are smaller than the 60-path budget of the
main comparison.
We therefore report the routes and TV of each complete accepted pool.
The route count $C$ corresponds to coverage $C/24$.
Coverage and TV depend on sample count, so we compare hardware results
with GPU sampling at the same budget.

For each fixed checkpoint, we ran 20 GPU sampling repetitions with the
same budget of 40 raw paths.
We tested both noiseless sampling and sampling with 0.5\% readout flips.
Each hardware row in \cref{tab:d3il-point-qpu} reports one run;
GPU standard deviations measure variation across the sampling repetitions.
\begin{table}[tbp]
\centering\normalsize
\setlength{\tabcolsep}{4pt}
\medmuskip=0mu
\caption{D3IL inference from the common stationary point. Each source uses 40 raw rollouts per prior. GPU rows give mean $\pm$ sample s.d. over 20 sampling repetitions of the same checkpoint; each QPU row is one 40-path run. All checkpoints use noisy parameter-shift training. Routes (out of 24) and TV are computed within geometrically valid paths.}
\label{tab:d3il-point-qpu}
\begin{tabular}{@{}llrr@{}}
\toprule
Prior & Sampling & \multicolumn{1}{c}{Routes $\uparrow$} & \multicolumn{1}{c}{TV $\downarrow$} \\
\midrule
Product-QCBM & GPU (0\% flips) & $9.45\,\pm\,1.82$ & $0.606\,\pm\,0.076$ \\
Product-QCBM & GPU (0.5\% flips) & $8.45\,\pm\,1.93$ & $0.648\,\pm\,0.081$ \\
\rowcolor{quantumrow}
Product-QCBM & Real QPU & $10.00$ & $0.583$ \\
\midrule
Entangled-QCBM & GPU (0\% flips) & $9.30\,\pm\,1.92$ & $0.613\,\pm\,0.080$ \\
Entangled-QCBM & GPU (0.5\% flips) & $9.40\,\pm\,1.85$ & $0.608\,\pm\,0.077$ \\
\rowcolor{quantumrow}
Entangled-QCBM & Real QPU & $13.00$ & $0.458$ \\
\bottomrule
\end{tabular}
\end{table}

Product-QCBM produced 13 accepted paths spanning 10 routes;
Entangled-QCBM produced 17 spanning 13 routes.
The noisy GPU references covered 5--13 and 6--14 routes, respectively.
Hardware coverage falls within these ranges, showing that the trained
priors can support complete recursive prediction on a QPU.

With $N\leq24$ accepted paths, each occupied route has frequency at
least $1/N\geq1/24$. TV against the uniform catalogue therefore equals
$1-C/24$, where $C$ counts occupied routes.
At this sample size, TV conveys the same information as route coverage.

We also compare the expected number of routes in a uniform eight-path
subset of each accepted pool (\cref{tab:d3il-point-qpu-eight}).
Pools with fewer than eight accepted paths are omitted only from this
supplementary statistic; their counts appear in the caption.
The hardware values are 6.92 and 7.05 routes, close after matching
the number of accepted paths.
\begin{table}[tbp]
\centering\small
\setlength{\tabcolsep}{4pt}
\medmuskip=0mu
\caption{D3IL route coverage standardized to eight accepted paths. Entries give the exact expected number of routes in an eight-path subset drawn uniformly without replacement from each saved valid pool. GPU means and sample s.d.s include only pools containing at least eight paths: 19/20 and 18/20 repetitions for Product-QCBM at 0\% and 0.5\% flips, and 19/20 and 20/20 for Entangled-QCBM. Each hardware entry describes one saved pool.}
\label{tab:d3il-point-qpu-eight}
\begin{tabular}{@{}lrrr@{}}
\toprule
Prior & \multicolumn{1}{c}{GPU (0\%)} & \multicolumn{1}{c}{GPU (0.5\%)} & \multicolumn{1}{c}{Real QPU} \\
\midrule
Product-QCBM & $6.42\,\pm\,0.54$ & $6.24\,\pm\,0.67$ & $6.92$ \\
Entangled-QCBM & $6.47\,\pm\,0.53$ & $6.18\,\pm\,0.53$ & $7.05$ \\
\bottomrule
\end{tabular}
\end{table}

\end{document}